\documentclass[10pt]{article}

\usepackage{fullpage}
\usepackage{setspace}
\usepackage{parskip}
\usepackage{titlesec}
\usepackage[section]{placeins}
\usepackage{xcolor}
\usepackage{breakcites}
\usepackage{lineno}
\usepackage{hyphenat}
\usepackage{subcaption}
\usepackage{natbib}
\usepackage{xfrac}
\usepackage{xurl} 

\usepackage{mathtools}
\DeclarePairedDelimiter{\ceil}{\lceil}{\rceil}

\PassOptionsToPackage{hyphens}{url}
\usepackage[colorlinks = true,
            linkcolor = blue,
            urlcolor  = blue,
            citecolor = blue,
            anchorcolor = blue]{hyperref}

\renewenvironment{abstract}
  {{\bfseries\noindent{\abstractname}\par\nobreak}\normalsize}
  {\bigskip}

\titlespacing{\section}{0pt}{*3}{*1}
\titlespacing{\subsection}{0pt}{*2}{*0.5}
\titlespacing{\subsubsection}{0pt}{*1.5}{0pt}

\usepackage{authblk}

\usepackage{tabularx} 
\usepackage{siunitx}
\usepackage{bm}

\usepackage{graphicx}
\usepackage[space]{grffile}
\usepackage{latexsym}
\usepackage{textcomp}
\usepackage{longtable}
\usepackage{tabulary}
\usepackage{booktabs,array,multirow}
\usepackage{amsfonts,amsmath,amssymb}
\providecommand\citet{\cite}
\providecommand\citep{\cite}

\newif\iflatexml\latexmlfalse

\AtBeginDocument{\DeclareGraphicsExtensions{.pdf,.PDF,.eps,.EPS,.png,.PNG,.tif,.TIF,.jpg,.JPG,.jpeg,.JPEG}}

\usepackage[utf8]{inputenc}
\usepackage[english]{babel}

\usepackage{moreverb} 

\immediate\write18{texcount -inc -incbib 
-sum paper.tex > /tmp/wordcount.tex}

\immediate\write18{texcount -char -freq
 paper.tex > /tmp/charcount.tex}

\usepackage{float}

\begin{document}

\title{CISO: Species Distribution Modeling Conditioned on Incomplete Species Observations}

\author[1]{Hager Radi Abdelwahed$^*$}
\author[1, 2]{Mélisande Teng$^*$}
\author[4]{Robin Zbinden$^*$}
\author[3]{Laura Pollock}
\author[1, 2]{Hugo Larochelle}
\author[4]{Devis Tuia}
\author[1,3]{David Rolnick}

\affil[1]{Mila--Quebec AI Institute, Montréal, Canada}
\affil[2]{Université de Montréal, Montréal, Canada}
\affil[3]{McGill University, Montréal, Canada}
\affil[4]{Environmental Computational Science and Earth Observation Laboratory, École Polytechnique Fédérale de Lausanne (EPFL), Lausanne, Switzerland}%
\renewcommand\Affilfont{\small}

\makeatletter
\renewcommand{\@date}{} 
\makeatother

\def\thefootnote{*}
\footnotetext{These authors contributed equally to this work. \\ \\ \textit{\normalsize Preprint, under review}}
\renewcommand{\thefootnote}{\arabic{footnote}}

\begingroup
\let\center\flushleft
\let\endcenter\endflushleft
\maketitle
\endgroup

\vspace{-2em}  
{\small
\noindent Correspondence: \texttt{hager.radi@mila.quebec}, \texttt{tengmeli@mila.quebec}, \texttt{robin.zbinden@epfl.ch}
}

\vspace{1.5em}


\selectlanguage{english}
\abstract{ 
Species distribution models (SDMs) are widely used to predict species' geographic distributions, serving as critical tools for ecological research and conservation planning. Typically, SDMs relate species occurrences to environmental variables representing abiotic factors, such as temperature, precipitation, and soil properties. However, species distributions are also strongly influenced by biotic interactions with other species, which are often overlooked. While some methods partially address this limitation by incorporating biotic interactions, they often assume symmetrical pairwise relationships between species and require consistent co-occurrence data. In practice, species observations are sparse, and the availability of information about the presence or absence of other species varies significantly across locations. To address these challenges, we propose CISO, a deep learning-based method for species distribution modeling Conditioned on Incomplete Species Observations. CISO enables predictions to be conditioned on a flexible number of species observations alongside environmental variables, accommodating the variability and incompleteness of available biotic data. We demonstrate our approach using three datasets representing different species groups: sPlotOpen for plants, SatBird for birds, and a new dataset, SatButterfly, for butterflies. Our results show that including partial biotic information improves predictive performance on spatially separate test sets. When conditioned on a subset of species within the same dataset, CISO outperforms alternative methods in predicting the distribution of the remaining species. Furthermore, we show that combining observations from multiple datasets can improve performance. CISO is a promising ecological tool, capable of incorporating incomplete biotic information and identifying potential interactions between species from disparate taxa.}

All datasets and model checkpoints are available at: \url{https://huggingface.co/cisosdm}. 
The source code is provided anonymously at: \url{https://anonymous.4open.science/r/SDMPartialLabels-B2EC}.

\onehalfspacing 

\section{Introduction}

Species distribution models (SDMs) are essential statistical tools widely used to monitor biodiversity and inform various ecological applications \citep{guisan2005predicting, jetz2019essential}, including understanding species-environment relationships \citep{dormann2012correlation}, guiding conservation strategies \citep{guisan2013predicting}, preventing the spread of invasive species \citep{barbet2018can}, and uncovering potential species interactions through co-occurrence patterns \citep{pellissier2010species, zurell2018joint}. 
Traditionally, SDMs correlate species occurrences with environmental factors such as climate and soil properties \citep{fourcade2018paintings}, helping to define the ecological niches of species and project these niches across geographic space to map species distributions \citep{elith2009species}. Key abiotic drivers, such as temperature, precipitation, and soil characteristics, often play a central role in determining habitat suitability \citep{pearson2003predicting, mod2016we}.

While abiotic factors are fundamental to species distribution, biotic interactions--such as competition, predation, and mutualism--can significantly influence species' ability to thrive and persist within particular environments \citep{wisz2013role, poggiato2021interpretations}. Incorporating biotic variables, such as the presence or absence of other species, into SDMs has been shown to substantially improve predictive accuracy and provide a more meaningful representation of the underlying ecological processes \citep{araujo2007importance, pellissier2010species, poggiato2025integrating}. For example, accounting for biotic interactions can help assess the likelihood of species invasions in the presence of native species \citep{poggiato2021interpretations}.

However, despite their acknowledged importance, biotic variables are rarely used as predictors in SDMs \citep{wisz2013role}. One major reason is the lack of detailed knowledge about interspecies interactions, a gap known as the \textit{Eltonian shortfall} \citep{hortal2015seven, pollock2025harnessing}. The specific relationships with other organisms are unknown for many species, making it difficult to determine \textit{a priori} the appropriate species to include as predictors. Moreover, even when biotic data are available, they are often incomplete or geographically limited, restricting the ability to include a comprehensive set of species interactions in SDMs. As a result, usually only species with well-documented distributions are incorporated, significantly limiting the inclusion of many species that could influence the distribution of the target species. This limitation greatly reduces the ability to account for the full range of biotic interactions. 

Most approaches incorporating species interactions rely heavily on co-occurrence data, i.e., samples where species have been observed at the same location, and ideally during the same time frame. Co-occurrences play a fundamental role in community ecology studies, where they are often analyzed with statistical methods to infer potential interactions \citep{dormann2018biotic}. Nevertheless, it has been emphasized that co-occurrences alone are usually not sufficient to provide evidence of species interactions, as shared habitat preferences or common evolutionary histories can also explain these patterns \citep{blanchet2020co}.

A popular approach for modeling species co-occurrences is offered by joint species distribution models (JSDMs), which aim to capture the relationships and dependencies among multiple species within a given environment \citep{pollock2014understanding, ovaskainen2016using, poggiato2021interpretations, wilkinson2021defining}. These models typically require complete presence and absence data for all species across surveyed sites, making them well-suited for controlled studies with comprehensive species inventories.
However, JSDMs struggle with incomplete species records, where presence or absence data for some species is missing \citep{romera2025should}. Such incompleteness arises for several reasons. First, different surveys may report different sets of species, making it challenging to integrate data from multiple sources without introducing assumptions about unrecorded species. Second, species observation checklists, such as those from eBird, are frequently incomplete, lacking data for some species. Given the high quantity of such records, it is crucial to develop methods that can effectively incorporate them rather than discard potentially valuable information. Finally, the rise of opportunistic observations generated by citizen science initiatives adds another layer of complexity \citep{escamilla2022taxonomic, cole2023spatial, romera2024should}. These datasets are often presence-only, recording observed species but providing no information on absences. 

This challenge extends beyond JSDMs to all existing SDMs that use biotic variables as input. Missing data is typically addressed by imputing pseudo-absences—introducing biases \citep{zbinden2024selection}—or discarding incomplete samples, resulting in the loss of valuable information. Additionally, JSDMs often assume symmetric pairwise relationships between species, an assumption that oversimplifies ecological dynamics, where interactions are frequently asymmetric and involve multiple species \citep{tikhonov2017using, blanchet2020co}. A more flexible SDMs framework is needed to effectively integrate available co-occurrence data and known absences while accounting for the complexity of species interactions.

We present a novel approach for SDMs, called CISO (for \textbf{C}onditioning on \textbf{I}ncomplete \textbf{S}pecies \textbf{O}bservations), that addresses these limitations and allows us to include a flexible number of species as input to the model. Our model leverages co-occurrence patterns and dynamically adjusts its predictions based on the presence or absence of other species. This is achieved using a deep learning approach designed to capture complex species interactions with environmental variables. Deep learning, with its capacity to process large amounts of unstructured data, is increasingly recognized as a powerful tool for modeling ecological processes and supporting wildlife conservation \citep{tuia2022perspectives, borowiec2022deep}. It has already proven successful in species identification from images and sounds \citep{norouzzadeh2021deep, kahl2021birdnet}, animal behavior study \citep{fazzari2024animal}, and ecological population monitoring \citep{joseph2020neural}.

Deep learning applications in SDMs are still in the early stages 
\citep{zbinden2023exploring, teng2023satbird, kellenberger2024performance, pollock2025harnessing}, but they have already demonstrated significant potential. Notably, deep learning can leverage large-scale crowdsourced datasets \citep{davis2023deep, lange2024active, brun2024multispecies}, capture non-linear relationships between environmental variables and realized abundance \citep{botella2018deep}, and integrate multimodal data beyond traditional environmental variables, such as satellite imagery and textual descriptions of range or habitat \citep{cole2020geolifedata, teng2023satbird, hamilton2024combining}. It can also support the flexible integration of spatial data at different scales and handle missing environmental input variables \citep{tiel2025multi, zbinden2025masksdm}. Additionally, deep learning offers the advantage of scalable modeling across multiple species simultaneously \citep{cole2023spatial}.

In this work, we adapt the C-Tran model \citep{cTran}, originally developed for multi-label image classification, where class labels are predicted based on the conditional dependencies of other known labels. We adapt this approach to SDMs, using available observations of other species, whether present or absent, to inform predictions of the target species distribution. Our model is therefore conditioned on both environmental factors (i.e., abiotic variables) and the partial occurrences of other species (i.e., biotic variables) at the location of interest. Since the species response to biotic variables varies with the values of the abiotic variables \citep{anderson2017and}, we employ a transformer model \citep{vaswani2017attention} that captures complex interactions between abiotic and biotic variables.

We demonstrate CISO on three datasets representing different groups of species: two publicly available datasets, sPlotOpen \citep{sabatini2021splotopen} and SatBird \citep{teng2023satbird}, which contain observations of plants and birds respectively, and a newly introduced dataset, SatButterfly, featuring butterfly observations across the USA. Our results show that species distribution modeling can be significantly improved by conditioning the model on the presence and absence of other species within the same dataset, yielding higher performance gains compared to alternative approaches. We provide qualitative examples illustrating how CISO can help identify species interactions, showcasing its potential for advancing ecological insights. Additionally, we explore the capability of combining multiple datasets, demonstrating that CISO can better predict species distributions in one dataset by conditioning on any available species information. This highlights the potential of CISO as a flexible and powerful approach for incorporating partial species observations. It represents an important step forward toward models that can integrate diverse datasets and account for multiple ecological processes simultaneously.

\section{Material and Methods}
\subsection{Datasets}

\begin{figure}
    \centering
    \includegraphics[width=0.97\linewidth]{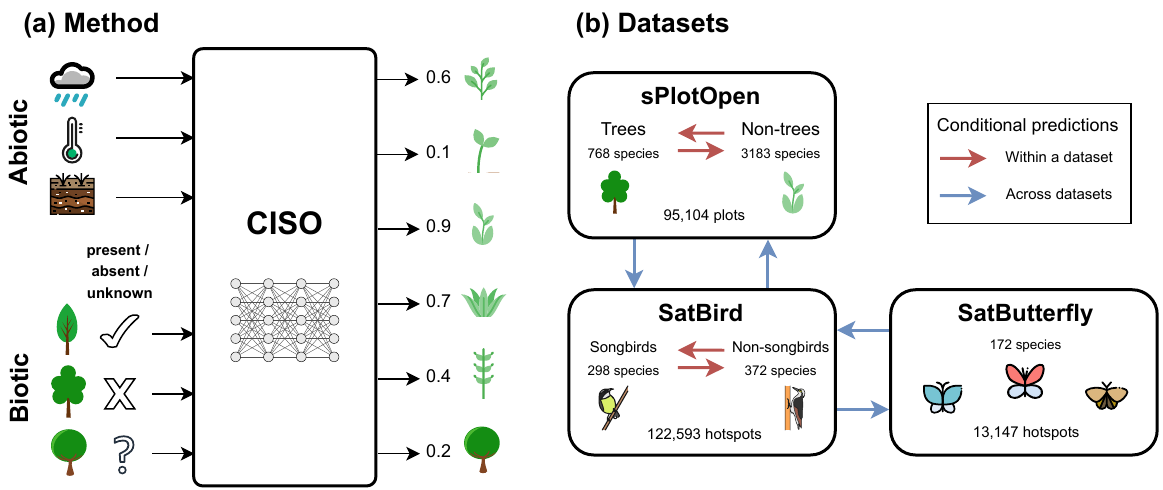}
    \caption{\textbf{(a)} Overview of our proposed approach, CISO, to condition SDMs on incomplete species observations: the model takes abiotic features and available biotic features as input and predicts species presence. CISO captures complex, non-linear relationships between species observations and both abiotic and biotic variables. \textbf{(b)} Summary of the three datasets in use:  sPlotOpen, Satbird, and SatButterfly, which we introduce in this work, along with the different setups for conditional inference, represented by the red and blue arrows.}
    \label{fig:datasets}
\end{figure}

We consider three distinct datasets of species occurrences, as illustrated in Figure \ref{fig:datasets}: \textit{sPlotOpen}, \textit{SatBird}, \textit{SatButterfly}. These datasets cover three taxonomic groups--plants, birds, and butterflies--that interact in various ecological contexts. We introduce the SatButterfly dataset in this paper and provide detailed descriptions of each dataset below.

We employ the same set of environmental abiotic predictors in all datasets. Climate conditions are captured using the $19$ bioclimatic variables from WorldClim \citep{oldworldclim}, which are widely applied in SDMs \citep{fourcade2018paintings}. Additionally, we include $8$ soil-related variables from SoilGrids \citep{hengl2017soilgrids250m}, which are particularly important to model plant species distributions \citep{austin2011improving, mod2016we}. These variables are provided as raster layers with spatial resolutions of approximately 1 km (climate) and 250 m (soil). For each observation, we extract the raster value at its geographic coordinates, producing a tabular data representation. This approach excludes broader spatial context, which has limited benefit at these coarse resolutions \citep{kellenberger2024performance, tiel2025multi}, and enables more lightweight models by reducing the number of parameters required for feature extraction. Although satellite imagery is available for both the SatBird and SatButterfly datasets, we do not include it in this study. Indeed, SDMs using environmental covariates have been shown to be strong baselines, and adding satellite imagery information does not consistently lead to performance gains that would justify extensive data downloads to create large-scale distribution maps \citep{tiel2025multi, picek2024geoplant} 


\subsubsection{sPlotOpen}

We utilize the sPlotOpen dataset \citep{sabatini2021splotopen}, a globally distributed, open-access collection of \num{95104} vegetation plots covering vascular plant species. sPlotOpen is an environmentally balanced subset of the larger sPlot dataset \citep{bruelheide2019splot}, specifically curated to mitigate the geographic biases present in the full dataset. For each plot, the presence or absence of every species is reported, resulting in a binary vector that indicates species occurrence. This plot-based data is particularly valuable for examining local co-occurrences and interactions among individual plant species.

We observe that species names are sometimes spelled differently within sPlotOpen (e.g., \textit{Calamagrostis epigeios} and \textit{Calamagrostis epigejos}). To address this, we use fuzzy string matching (using the \texttt{thefuzz\footnote{\url{https://github.com/seatgeek/thefuzz}}} python package) to compare species names and merge pairs with a similarity score greater than \num{90}. We find 11 pairs of species to be merged, which are listed in Appendix \ref{appendix:splotopen}.

Additionally, the distribution of presence observations is highly imbalanced, with many species recorded at only a few locations, while a relatively small number of species are observed much more frequently. To reduce computational overhead and ensure sufficient data for robust analysis, we retain species with at least \num{100} presence observations across all locations, resulting in \num{3951} distinct species. On average, there are \num{20} species present per plot. These co-occurrences are crucial for our model to capture dependencies between species.
To account for spatial auto-correlation between the plots, we partition the data into training, validation, and test sets using spatial block-cross validation \citep{roberts2017cross}. A visualization of these splits is provided in Appendix \ref{appendix:splotopen}.

For our within-sPlotOpen experiments, we focus on predicting non-tree species based on observations of tree species, and \textit{vice versa}. Practical considerations drive this distinction: while remote sensing imagery often provides reliable data on tree species, accessing specific ground sites to gather information on understory or small non-tree species can be more challenging. Additionally, tree species are typically well-documented and easier to observe \citep{cazzolla2022number}. Moreover, we expect strong ecological interactions between these two groups. For instance, trees often act as nurse plants, facilitating the establishment and survival of non-tree species in harsh environments \citep{padilla2006role, gomez2009role}. Trees are a polyphyletic group with some variation in definitions; we adopt the definition of the Botanic Gardens Conservation International (BCGI) database\footnote{\url{https://tools.bgci.org/global_tree_search.php}}, which lists \num{57681} tree species from around the world. Using this definition, the sPlotOpen species under consideration comprise \num{767} tree species and \num{3183} non-tree species.

\subsubsection{SatBird}
The SatBird dataset \citep{teng2023satbird} was proposed for the task of jointly predicting bird species encounter rates based on satellite images and environmental variables. The encounter rate is defined as the frequency with which a species has been observed at a specific location, derived from complete checklist records from eBird \citep{eBird:HCLN}. This rate estimates the probability that a visitor will observe a species during their visit, with values ranging from \num{0} to \num{1}.
In our study, we use the USA-summer subset of SatBird which encompasses \num{122593} locations across the USA and includes data on \num{670} bird species. We use the same train, validation, and test sets of \cite{teng2023satbird}.

Similarly to sPlotOpen, we categorize the species into two groups: \num{372} non-songbird species and \num{298} songbird species, to predict non-songbirds based on observation of songbirds, and \textit{vice versa}. Unlike trees, songbirds represent a monophyletic group (order Passeriformes), often characterized by their diverse and elaborate vocalizations, which make them more recognizable and detectable.
Songbirds are a species-diverse group found across a wide variety of habitats and occupying a broad range of ecological niches, making them an effective taxon for predictive analysis. Motivating our analysis, interspecific competition among bird species has been identified as an important factor influencing populations \citep{dhondt2012interspecific}, and incorporating biotic variables has proven beneficial for modeling bird species distributions in SDMs \citep{zurell2017integrating}.

\subsubsection{SatButterfly} 
\label{sec:satbutterfly_dataset}

We prepare and release the SatButterfly dataset, modeled after the SatBird dataset and derived from eButterfly checklists \citep{prudic2017ebutterfly}. 
Following the exact methodology of \citep{teng2023satbird}, this dataset includes \num{13147} locations across the continental USA, capturing encounter rates of \num{601} butterfly species. Unlike SatBird, SatButterfly includes observations collected year-round, as most butterflies are non-migratory \citep{chowdhury2021migration}. The checklists span observations from 2010 to 2023. Satellite imagery is available for each location, although we do not use it in this work. Similar to sPlotOpen, we focus on species with at least \num{100} recorded occurrences, resulting in a total of \num{172} butterfly species. However, unlike sPlotOpen and SatBird, we do not sub-categorize species in SatButterfly for conditional predictions due to the relatively lower number of observations compared to the other datasets. More details about the dataset are provided in Appendix \ref{appendix_satbutterfly_dataset}.

\subsubsection{Combining Datasets} \label{sec:combine_datasets}

Significant taxonomic biases in data collection result in certain species groups being far more frequently surveyed than others. For example, bird observations are relatively abundant in some regions, primarily due to the extensive efforts of birdwatchers contributing data through platforms such as eBird. In contrast, butterfly observations are further behind in terms of geographic coverage and volume of observations, although data availability is improving through initiatives such as eButterfly \citep{prudic2017ebutterfly}, inspired by eBird.

We propose leveraging well-documented species groups to improve the modeling of data-deficient species. To this end, we combine different datasets and align observations recorded at the same location, which we refer to as \textit{co-locations}. Note that a co-location does not necessarily imply a co-occurrence, as the data may have been collected at different times. The implications and limitations of this approach are further discussed in Section \ref{sec:discussion}.

We identify co-locations between dataset \textit{A} and dataset \textit{B} using the following approach. First, we perform BallTree-based clustering \citep{omohundro1989five} with locations of dataset \textit{A} designated as centroids. For each centroid, we search within a $1$ km radius for the nearest location from dataset \textit{B} using the Haversine distance. If a neighboring location from dataset \textit{B} is found within this radius, we assign its species data to the corresponding location in dataset \textit{A} (if there are multiple such locations, we assign it to the closest). This process yields a subset of dataset \textit{A}’s training, validation, and test locations that also contain species data from dataset \textit{B}, thus defining the co-locations between dataset \textit{A} and dataset \textit{B}. Note that this procedure is asymmetric.

We apply this clustering approach, where we use the SatBird USA-summer subset as dataset \textit{A} and SatButterfly as dataset \textit{B}. We obtain \num{6684} co-locations that follow SatBird training, validation, and test splits, which can be used to study the ecological interactions between butterfly species and bird species. We hypothesize that incorporating bird encounter rates into species distribution models will enhance predictions of butterfly encounter rates, and \textit{vice versa}, by capturing shared biotic and abiotic factors that may be inferred from species occurrences. For example, the Karner blue (a butterfly) and Kirtland's warbler (a songbird) are rare species that both inhabit pine barrens ecosystems and thus frequently co-occur \citep{wood2011effects}. 
For training, we include all locations in the SatBird and SatButterfly training sets, while the validation and test sets are restricted to co-located data, ensuring that evaluation is performed with complete information.

We follow the same clustering approach for combining SatBird and sPlotOpen. We designate the SatBird USA-summer subset as dataset \textit{A} and sPlotOpen as dataset \textit{B}, resulting in \num{1161} co-locations. Predicting bird distributions based on plant observations, and \textit{vice versa}, is particularly relevant, as these groups are ecologically interconnected. Birds frequently rely on specific trees for nesting \citep{cockle2011woodpeckers} and play a key role in seed dispersal \citep{wenny2001advantages}. When combining the datasets, we only consider locations within the continental USA, as this is the region covered by the SatBird dataset. Consequently, the sPlotOpen dataset is also limited to this region, resulting in 607 plant species being considered. To ensure that the evaluation is based on complete information, both the validation and test sets contain only the co-located observations.

\subsection{The CISO approach}\label{section:CISO}

We aim at developing a multi-species distribution model that predicts suitability scores of a set of species $\mathcal{C}$ simultaneously, conditioned on both environmental variables $x$ and a subset of species $\mathcal{C}_{\text{known}} \subset \mathcal{C}$, which consists of any $k$ species whose presence or absence is already known. The size of this subset varies across locations depending on the available species data. Crucially, the model should perform well even when no additional species information is provided (i.e., when $k = 0$ and $\mathcal{C}_{\text{known}} = \emptyset$), relying solely on abiotic variables for predictions.
Our approach, CISO (Conditioning SDMs on Incomplete Species Observations), builds upon C-Tran \citep{cTran}, a method originally designed for multi-label image classification that incorporates known but incomplete class labels into its predictions. C-Tran uses partial label information to infer the presence or absence of other classes by first extracting image features and then modeling dependencies between these features and the known labels. This is accomplished through a label masking strategy during training, encoding three possible label states (positive, negative, or unknown) that are fed into the model. As a result, the model can leverage any arbitrary subset of available labels, in addition to the image itself, to inform its predictions.

\begin{figure}
    \centering
    \includegraphics[width=0.9\linewidth]{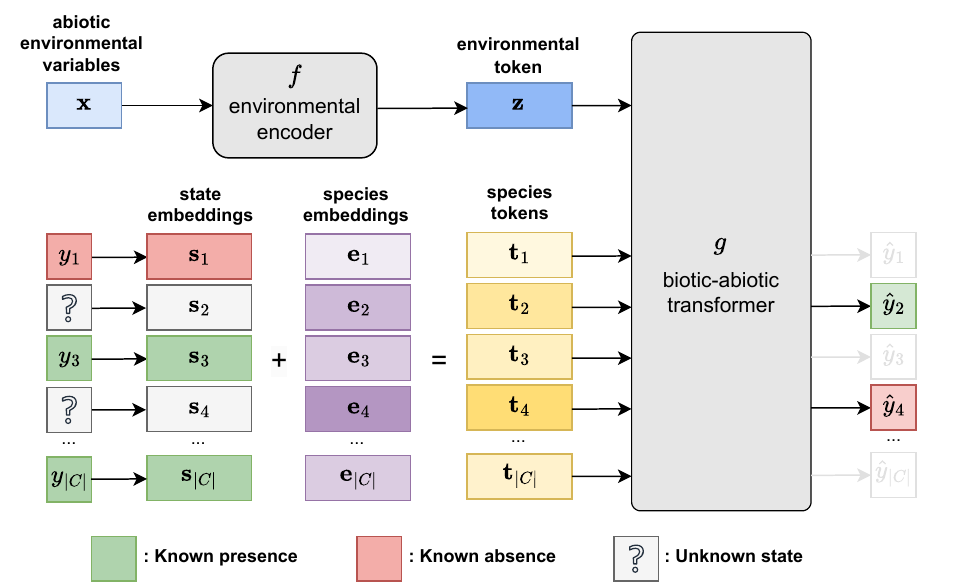}
    \caption{Figure shows the architecture of CISO. CISO incorporates incomplete species observations at inference. The first branch processes abiotic variables, while the second branch integrates the known presences or absences of other species.}
    \label{fig:architecture}
\end{figure}

We adapt this framework to SDMs, as illustrated in Figure \ref{fig:architecture}. Each species $c \in \mathcal{C}$ is represented by a vector $\bm{e}_c \in \mathbb{R}^d$, referred to as the \textit{species embedding}. These embeddings can be learned by the model or, if available, can encode species-specific information such as traits. Additionally, we define a \textit{state embedding} $\bm{s}_c \in \mathbb{R}^d$ to indicate whether species $c$ is known to be present, absent, or its status is unknown. Specifically, $\bm{s}_c$ is set to the vector $\bm{p}$ for presence, $\bm{a}$ for absence, or $\bm{u}$ for unknown. These state vectors are learned during training and represent their respective states. The state embedding is then added to the species embedding to form the \textit{species token}:
\begin{align*}
    \bm{t}_c = \bm{s}_c + \bm{e}_c.
\end{align*}
The vectors $\bm{e}_c$, $\bm{s}_c$, and $\bm{t}_c$ for all species $c$ are organized into the respective matrices $\bm{E}, \bm{S}, \bm{T} \in \mathbb{R}^{|\mathcal{C}|\times d}$.

The abiotic variables $\bm{x}$ are processed by a neural network $f(\cdot)$, which we term the \textit{environmental encoder}, designed to learn non-linear relationships among these environmental inputs. The encoder produces an environmental representation $\bm{z} = f(\bm{x}) \in \mathbb{R}^{d}$. This representation, along with the species tokens $\bm{T}$, is fed into a neural network $g(\cdot)$ with a transformer-based architecture \citep{vaswani2017attention}, that we call \textit{biotic-abiotic transformer}. Originally designed for processing textual data, the transformer has since been successfully applied to a variety of data types, including images and tabular data \citep{dosovitskiy2020image, gorishniy2021revisiting}. It utilizes attention mechanisms to selectively focus on different parts of the input, enabling it to capture interactions and dependencies effectively. 
In our context, the transformer models the interactions between environmental factors and species occurrences by learning relevant relationships through self-attention. Ultimately, the transformer outputs suitability scores $\bm{\hat{y}} = g(\bm{Z}, \bm{T})$ for all species in $\mathcal{C}$, reflecting the likelihood of suitable biotic and abiotic conditions for each species.

To ensure the model can handle any subset $\mathcal{C}_{\text{known}}$, we adapt its training procedure. C-Tran introduces \textit{Label Mask Training} (LMT), a method that trains the model to learn label correlations and generalize across different inference scenarios. Inspired by BERT's masked language model \citep{devlin2018bert}, LMT randomly masks a proportion of labels during training and uses the remaining labels to predict the masked ones. In our case, we randomly select a subset $\mathcal{C}_{\text{known}}$ and indicate their presence or absence via the state embedding $\textbf{s}$. At each training step, the number of known species $k$ is randomly chosen from the range $0$ to $\sfrac{3}{4}\cdot |\mathcal{C}|$, and $k$ species are drawn from $\mathcal{C}$. The $|\mathcal{C}| - k$ species are assigned a state value of \textit{unknown}. By exposing the model to varying amounts of known species data during training, it learns to handle a wide range of species combinations. 

During inference, any available observations of other species can be used to predict the suitability scores of target species at a given location. The subset of known species $\mathcal{C}_{\text{known}}$ is incorporated similarly to what is done during training, conditioning predictions on their presence or absence. When no species information is available, the model relies solely on environmental variables. This allows a direct comparison between predictions with and without conditioning on other species, showing how the presence or absence of one species influences the predicted suitability of another.

\subsubsection*{Encoding species encounter rates}

In C-Tran \citep{cTran}, the presence or absence of a label is represented by a binary signal (1 or 0, respectively) in the state embedding, which is also how the data in the sPlotOpen dataset is structured. We adapt this approach to support the SatBird and SatButterfly datasets, where species labels are represented by encounter rates at a given location. Unlike binary presence/absence labels, these encounter rates are continuous values between 0 and 1, providing more detailed information about species occurrence. To leverage this richer information, we encode the encounter rates into the state embeddings $\bm{s}$ using a \textit{discrete binning} approach. We discretize strictly positive values into $n_b$ equally spaced bins such that an encounter rate value $r$ will be assigned to the bin indexed by $\ceil{r \cdot n_b} / n_b$. Each bin corresponds to a unique state embedding, resulting in $n_b +2$ distinct embeddings, including the absence and unknown embeddings. In this way, the encounter rate is represented by the state embedding corresponding to its assigned bin. 
This generalizes the setting of C-Tran, where presence and absence can be interpreted as binary discrete binning.
Additionally, we explore continuous projection encoding, mapping the scalar encounter rate values directly into a higher-dimensional space. Following the method from \cite{gorishniy2022embeddings}, we apply a linear projection to the scalar value, followed by periodic activation functions. This allows the model to capture subtle variations in the encounter rates. Although our main experiments use discrete binning, we conduct ablation studies comparing this approach to continuous and linear projections of encounter rates following \cite{gorishniy2022embeddings}; we report results in Appendix \ref{appendix:ablation_continuous_projection}.

In all approaches, when we mask out the encounter rate, we use the learned \textit{unknown} embedding similarly to when handling a binary presence or absence.
Whenever we refer to incorporating species observations as presence or absence in CISO for SatBird and SatButterfly, we include encounter rate values rather than binary presence indicators.

\subsubsection*{Handling missing species observations}

One main challenge is to handle training datasets where the available species information is incomplete or missing for certain species. This issue is common, for example, in presence-only data sources, but also in eBird checklists, which are often incomplete due to incidental observations and historically partial lists \citep{johnston2021analytical}. It also arises when different datasets are combined, and include locations where only species from dataset \textit{A} or \textit{B} are available, at a time. In such cases, there may be a significant amount of missing data on species presence or absence. To address this, we represent missing observations using the \textit{unknown} state embedding and adapt the training procedure to handle the incomplete data. Specifically, during training, instead of sampling the $k$ species in $\mathcal{C}_{\text{known}}$ from the full set $\mathcal{C}$, we sample them from the set $\mathcal{C}_{\text{available}} \subset \mathcal{C}$, which includes the $l$ species for which data is available. Accordingly, $\mathcal{C}_{\text{known}} \subset \mathcal{C}_{\text{available}}$ with $k$ drawn between $0$ and $l$. This modification enables the model to fully leverage all available co-occurrence data during training.

\subsection{Experimental setup}
In this section, we describe the experimental setup used to evaluate CISO and the baseline approaches for comparison. Further details about the experiments can be found in Appendix \ref{appendix:exp_details}.

\subsubsection{Models}\label{baselines}

To better understand and disentangle the contributions of different components of our approach, we compare CISO with other baseline models, including traditional SDMs methods such as Maxent. In this section, we introduce the models considered in this study and highlight how each comparison helps assess specific aspects of our design, such as the effect of conditioning on available biotic information and the different ways this conditioning can be implemented.

\textbf{Linear model.} The first baseline is a linear model that takes environmental variables as input and predicts suitability scores for the $|\mathcal{C}|$ species. A sigmoid function is applied to the outputs to obtain scores between $0$ and $1$.

\textbf{Maxent.} We extend the linear model by first transforming the environmental variables using Maxent feature functions \citep{phillips2006maximum}. Specifically, we consider five types of features: linear, hinge, product, threshold, and quadratic, computed via the \texttt{elapid}\footnote{\url{https://earth-chris.github.io/elapid/}} Python package, resulting in \num{1161} features.

\textbf{Multilayer perceptron (MLP).} This baseline is designed to learn non-linear interactions among the environmental variables and predict suitability scores for all species. This model consists of \num{2} linear layers, each with \num{256} hidden neurons, interleaved with the ReLU activation function. A final linear layer produces suitability scores for the $|\mathcal{C}|$ species. The scores are constrained to the range $[0, 1]$ using a sigmoid function. This baseline provides a measure of how a basic neural network model performs when relying solely on environmental variables. An ablation study on the number of layers of the MLP baseline is conducted in Appendix \ref{appendix:ablation_num_layers}.

\textbf{MLP++.}
We modify the MLP baseline to incorporate partial observations of some species. This enhanced model, referred to as MLP++, retains the same architecture as the MLP baseline but includes additional inputs representing the state of each species: \textit{negative}, \textit{positive}, or \textit{unknown}. For the SatBird and SatButterfly datasets, the \textit{positive} state is represented using discrete bins, following CISO, as explained in \ref{section:CISO}. These additional state inputs are one-hot encoded and given to the model alongside the environmental predictors. This baseline provides the simplest approach for incorporating partial observations as input into a neural network.

\textbf{CISO.}
CISO reuses the MLP baseline as the backbone of its architecture for the environmental encoder $f(\cdot)$. However, the final linear layer is removed, and the resulting representation is passed to the biotic-abiotic transformer, along with the species tokens, obtained as described in Section \ref{section:CISO}. Following \cite{cTran}, the transformer comprises \num{3} self-attention layers, each with \num{4} attention heads. For the discrete binning of encounter rates in the SatBird and SatButterfly datasets, we use $n_b = 4$ bins to represent present species. 

We evaluate CISO and MLP++ by comparing their performance under two conditions: without partial species observations (unconditioned) and with partial species observations (conditioned). This comparison allows us to assess whether incorporating additional biotic interactions improves performance and whether CISO is an effective approach for leveraging this information.

\subsubsection{Training details}
We train CISO and all baseline models for \num{50} epochs using the AdamW optimizer \citep{adamW}. Across all datasets, we use the binary cross-entropy (BCE) loss, following \cite{teng2023satbird}:
\begin{align*}
    \mathcal{L}_{BCE}= -\sum_{c \in \mathcal{C} \setminus \mathcal{C}_{\text{known}}} y_c\log(\hat{y}_c) + (1-y_c)\log(1-\hat{y}_c),
\end{align*}
where $\hat{y}$ represents the model predictions, and $y$ the ground truth. The BCE loss is suited for the regression task of SatBird and SatButterfly, as the targets are probabilities in [0,1], and the BCE supports such values.
All baseline models, except for CISO, are trained on CPUs. While CISO models are also compatible with CPU training, we opt to use a single GPU for their training to accelerate the process. Inference for all models, including CISO, is performed on CPUs.

\subsubsection{Evaluation Metrics}
We use different metrics to evaluate the model's performance based on the type of data available. For species in the sPlotOpen dataset, where presence and absence data are available, we evaluate the baselines using the Area Under the Receiver Operating Characteristic Curve (AUC). The AUC is widely used for assessing SDMs and allows evaluation without binarizing the predictions. We compute the AUC only for species with a least one recorded presence in the test set.
For species in the SatBird and SatButterfly datasets, where ground truths are scalar encounter rates, we employ several metrics. Our primary focus is on the Mean Absolute Error (MAE) and the top-$k$ accuracy. The top-$k$ refers to the adaptive top-$k$ metric defined in the work of \cite{teng2023satbird}, where for each hotspot, $k$ corresponds to the number of species with non-zero ground truth encounter rates. Further details about model training and selection, as well as additional evaluation metrics, including the Mean Squared Error (MSE) and the fixed top-10 and top-30 accuracy, are presented in Appendix \ref{appendix:exp_details}.

\section{Results}

\subsection{Within-dataset}

\begin{table}
    \renewcommand{\arraystretch}{1.2}
    \centering
    \resizebox{\columnwidth}{!}{%
    \begin{tabular}{l|cc|cccc|}
         & \multicolumn{2}{c|}{sPlotOpen} & \multicolumn{4}{c|}{SatBird} \\
         & \multicolumn{2}{c|}{AUC (\%)} & \multicolumn{2}{c|}{Top-$k$ (\%)} & \multicolumn{2}{c|}{MAE [$\times 10^{2}$]} \\ \hline
        \multicolumn{1}{c|}{Model} & tree & non-tree & songbird & \multicolumn{1}{c|}{non-songbird} & songbird & non-songbird \\ \hline
        \textbf{Unconditioned} & & & & \multicolumn{1}{c|}{} & & \\
        Linear & $96.69 \pm 0.01$ & $93.74 \pm 0.00$ & $61.75 \pm 0.04$ & \multicolumn{1}{c|}{$52.26 \pm 0.08$} & $8.99 \pm 0.02$ & $7.66 \pm 0.01$ \\
        Maxent & $98.24 \pm 0.00$ & $96.17 \pm 0.00$ & $67.08 \pm 0.02$ & \multicolumn{1}{c|}{$57.52 \pm 0.04$} & $3.43 \pm 0.01$ & $1.39 \pm 0.01$ \\
        MLP & $98.39 \pm 0.03$ & $96.58 \pm 0.01$ & $67.99 \pm 0.04$ & \multicolumn{1}{c|}{$58.15 \pm 0.09$} & $3.34 \pm 0.02$ & $1.37 \pm 0.01$ \\
        MLP++ & $98.15 \pm 0.04$ & $96.11 \pm 0.03$ & $65.81 \pm 0.19$ & \multicolumn{1}{c|}{$56.76 \pm 0.14$} & $3.49 \pm 0.06$ & $1.37 \pm 0.03$ \\
        CISO & $98.49 \pm 0.05$ & $96.49 \pm 0.07$ & $69.30 \pm 0.21$ & \multicolumn{1}{c|}{$59.41 \pm 0.25$} & $3.19 \pm 0.09$ & $1.28 \pm 0.04$ \\ \hline
        \textbf{Conditioned} & & & & \multicolumn{1}{c|}{} & & \\
        MLP++ & $98.82 \pm 0.03$ & $96.92 \pm 0.02$ & $69.98 \pm 0.30$ & \multicolumn{1}{c|}{$61.76 \pm 0.27$} & $2.75 \pm 0.02$ & $1.06 \pm 0.02$ \\
        CISO & $\mathbf{99.12} \pm 0.04$ & $\mathbf{97.39} \pm 0.14$ & $\mathbf{72.86} \pm 0.37$ & \multicolumn{1}{c|}{$\mathbf{63.56} \pm 0.49$} & $\mathbf{2.57} \pm 0.08$ & $\mathbf{1.04} \pm 0.03$
    \end{tabular}
}
    \setlength{\abovecaptionskip}{10pt}
    \caption{\textbf{Within-dataset}: Performance comparison of the different approaches applied to different species groups within the same dataset. Models with \textit{conditioned} inference incorporate partial information about the presence or absence of species from the other group. Bolded scores indicate the best performance for the given metric and dataset. Performance metrics are averaged over three random seeds, with the standard deviations reported.}
    \label{tab:within_dataset}
\end{table}

We first evaluate our approach in scenarios where biotic variables consist of observations from species within the same dataset. The results are presented in Table \ref{tab:within_dataset}. For the sPlotOpen dataset, we predict tree species conditioned on observations of non-tree species and \textit{vice versa}. Similarly, for the SatBird dataset, we predict songbirds conditioned on observations of non-songbird species and \textit{vice versa}. We also evaluate all methods in an unconditioned setting, where no partial observations are provided as input; indeed, for the linear, Maxent, and MLP models, this is the only possible setting due to the inability of these models to represent conditioning on partial observations without additional modifications.

First, we observe that the linear model is significantly outperformed by all other methods, indicating that non-linearity is essential for achieving optimal performance.
In the unconditioned setting, Maxent and the multilayer perceptron baselines, MLP and MLP++, perform similarly, while CISO achieves slightly better results.
However, when conditioning on partial observations, both MLP++ and CISO show significant performance gains across all datasets and metrics, particularly on SatBird. This highlights the benefit of integrating biotic variables as inputs.
Overall, CISO achieves the highest performance, demonstrating its effectiveness in handling incomplete species observations.

\begin{figure}
    \centering
    \includegraphics[width=1\linewidth]{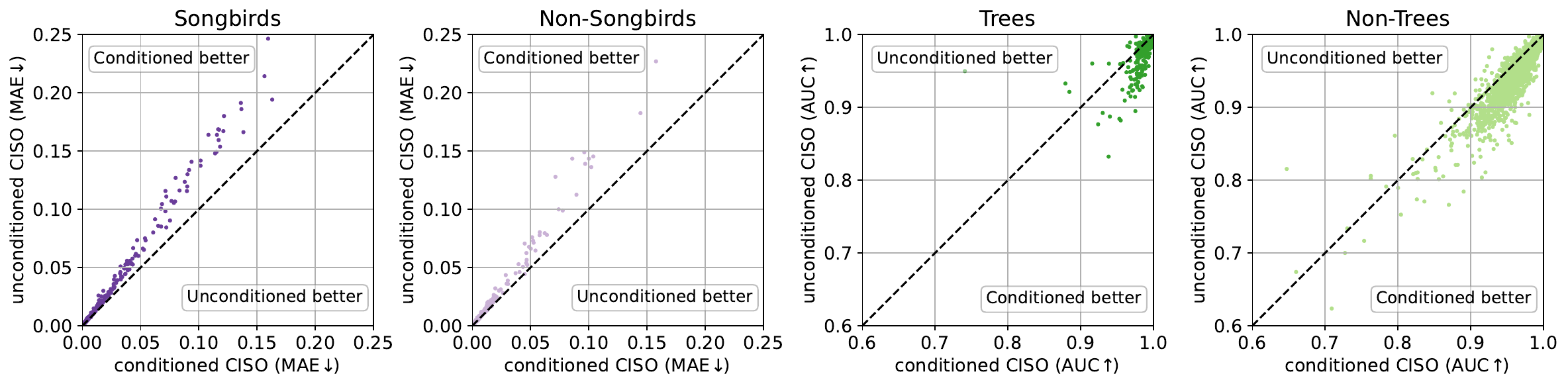}
    \caption{Comparison of species performance between the conditioned and unconditioned CISO models. Each dot represents a species, with its position indicating the performance of the unconditioned and conditioned models. The dashed line indicates equal performance between the two models. The title of each plot corresponds to the subgroup of species predicted, conditioned on, or not conditioned on the rest of the species of the dataset (SatBird in purple and sPlotOpen in green). Small MAE is better, while high AUC is better.}
    \label{fig:species_distr_perf}
\end{figure}

Conditioning on observations from other species improves performance, but does it uniformly benefit all species? In Figure \ref{fig:species_distr_perf}, we show the performance of both the unconditioned and conditioned CISO models for each species. For the species in SatBird (Songbirds and Non-songbirds), we observe a consistent relative improvement across all species, with a larger gain seen for species with lower unconditioned performance. For sPlotOpen, a clear trend in favor of conditioning is also observed for most, though not all, species, but the effect is not as uniform. This more complicated behavior could be due to the higher number of species considered, which increases the likelihood of the model learning spurious correlations in species co-occurrence patterns that negatively impact some predictions.

\subsection{Across Datasets}
We also evaluate our approach in a more challenging scenario, where biotic interactions are considered across disparate taxa using observations from separate datasets. The results are presented in Table \ref{tab:across_dataset}. Specifically, we consider predicting SatBird bird species from observations of butterfly species from SatButterfly, and \textit{vice versa}. Similarly, we predict plant species from sPlotOpen based on bird species observations from SatBird, and \textit{vice versa}. Performance is evaluated on co-locations between the two datasets.

\begin{table}
    \renewcommand{\arraystretch}{1.1}
    \centering
    \resizebox{0.7\columnwidth}{!}{
        \begin{tabular}{l|cc|cc|}
             & \multicolumn{2}{c|}{sPlotOpen $\leftrightarrow$ SatBird} & \multicolumn{2}{c|}{SatBird $\leftrightarrow$ SatButterfly} \\
             & \multicolumn{2}{c|}{1161 co-locations}&  \multicolumn{2}{c|}{6684 co-locations}\\ 
             & \multicolumn{1}{c}{AUC (\%)} & \multicolumn{1}{c|}{Top-$k$ (\%)} & \multicolumn{1}{c}{Top-$k$ (\%)} & \multicolumn{1}{c|}{Top-$k$ (\%)} \\ \hline
            \multicolumn{1}{c|}{Model} & sPlotOpen & SatBird & SatBird & \multicolumn{1}{c|}{SatButterfly} \\ \hline
            \textbf{Unconditioned} & & & & \\
            Linear & $82.69 \pm 0.03$ & $48.87\pm 0.07$ & $62.97 \pm 0.38$ & \multicolumn{1}{c|}{$29.56 \pm 0.11$} \\
            Maxent & $94.97 \pm 0.24$ & $58.69 \pm 0.24$ & $70.06 \pm 0.28$ & \multicolumn{1}{c|}{$32.90 \pm 0.38$} \\
            MLP & $95.28 \pm 0.03$ & $60.99 \pm 0.21$ & $71.12 \pm 0.11$ & \multicolumn{1}{c|}{$33.30 \pm 0.56$} \\
            MLP++ & $94.27 \pm 0.15$ & $58.17 \pm 0.18$ & $69.04 \pm 0.28$ & \multicolumn{1}{c|}{$32.15 \pm 0.36$} \\
            CISO & $\mathbf{95.49} \pm 0.24$ & $\mathbf{62.78} \pm 0.03$ & $72.42 \pm 0.02$ & \multicolumn{1}{c|}{$33.59 \pm 0.18$} \\ \hline
            \textbf{Conditioned} & & & & \\
            MLP++ & $92.24 \pm 0.63$ & $53.86 \pm 0.34$ & $67.96\pm 0.32$ & \multicolumn{1}{c|}{$30.83 \pm 0.55$} \\
            CISO & $94.79 \pm 0.24$ & $59.43 \pm 0.03$ & $\mathbf{72.43} \pm 0.17$ & \multicolumn{1}{c|}{$\mathbf{34.49} \pm 0.34$}
        \end{tabular}
    }
    \setlength{\abovecaptionskip}{10pt}
    \caption{\textbf{Across Datasets}: Performance comparison of the different approaches on different datasets. Conditioned approaches integrate observations from another dataset of different species. Bolded scores indicate the best performance for the given metric and dataset. Performance metrics are averaged over three random seeds, with standard deviations reported.}
    \label{tab:across_dataset}
\end{table}

In both tasks, we find that the linear model is again outperformed by all other methods. Across both unconditioned and conditioned settings, MLP++ does not outperform MLP.  
In contrast, CISO surpasses MLP and MLP++, showing the potential of this architecture for modeling species distribution.
Notably, we observe that the performance of MLP++ and CISO is generally lower or similar in the conditioned setting compared to the unconditioned case, highlighting the increased difficulty of this task. The only exception is the task of predicting butterfly encounter rates from bird species observations, where CISO in the conditioned setting achieves the best results. Unlike the within-dataset setup, in which information on all species is available at every location, fewer co-locations are available in the across-datasets scenario due to the combination of data from different sources. We discuss this further in Section \ref{sec:discussion}.

\subsection{Qualitative analysis}
\begin{figure}
    \centering
    \includegraphics[width=1\linewidth]{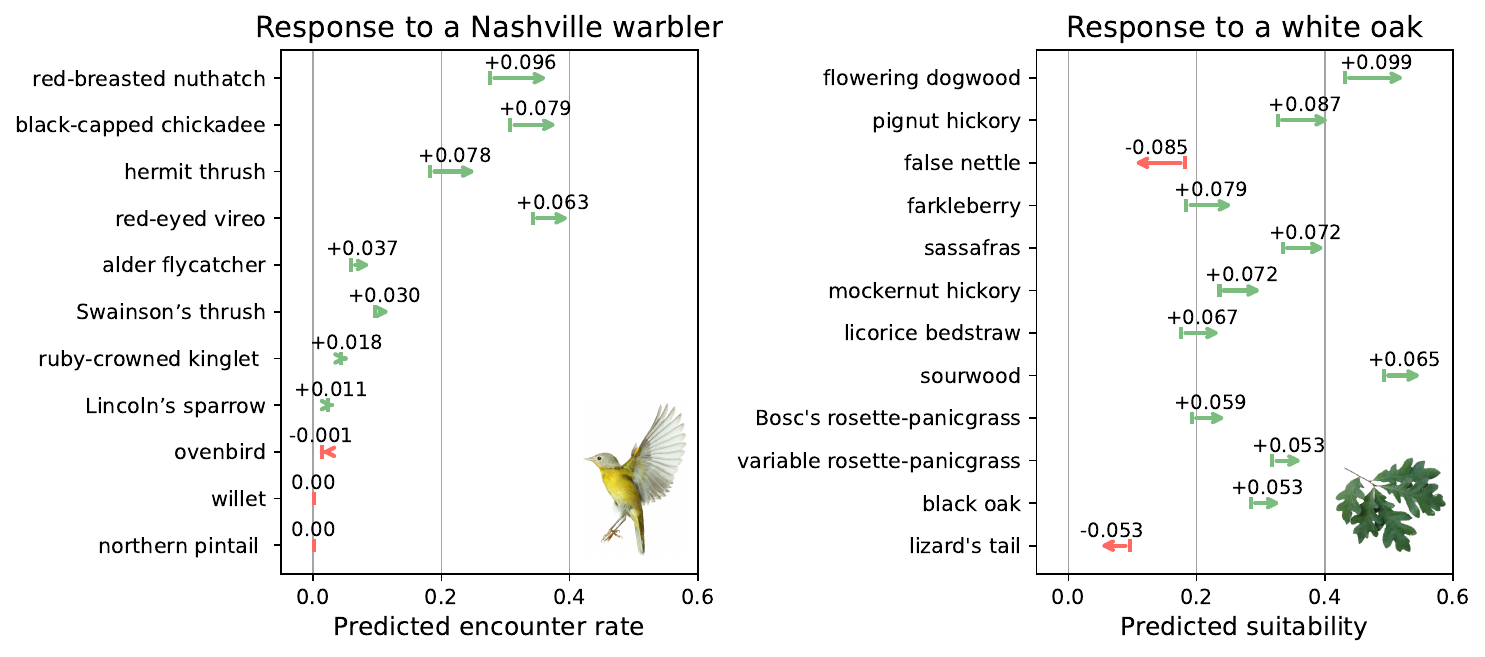}
    \caption{Change in CISO model predictions when conditioned on the presence of another species. Each arrow originates from the average prediction of the unconditioned CISO model for the test set locations of the respective dataset and ends at the average prediction when CISO is conditioned on the presence of the other species. 
    \textbf{Left:} Change in predicted encounter rates for certain species in SatBird when CISO is conditioned on the (true) presence of the Nashville warbler (\textit{Leiothlypis ruficapilla}). \textbf{Right:} Change in predicted suitability for certain species in sPlotOpen when CISO is conditioned on the (true) presence of the white oak (\textit{Quercus alba}).}
    \label{fig:species_response}
\end{figure}

We analyze how CISO's predictions change when conditioned on the presence of another species, as shown in Figure \ref{fig:species_response}. The left panel illustrates how the model's response varies for different bird species in the SatBird dataset when given information about a single species, in this case, the Nashville warbler (\textit{Leiothlypis ruficapilla}). We examine the same set of species as in \cite{harris2015generating}. The figure highlights the change in predicted encounter rates when CISO is conditioned on the ground truth encounter rate of the Nashville warbler, compared to when no species information is provided. The changes are averaged over test set locations where the Nashville warbler has a positive encounter rate. We observe moderate changes that generally align with the findings of \cite{harris2015generating}, except for the ovenbird. Notably, the red-breasted nuthatch shows a positive response to the presence of the Nashville warbler. These species have similar habitat preferences, with the Nashville warbler breeding in Northern forests that overlap with the habitat of the red-breasted nuthatch.

The right panel of Figure \ref{fig:species_response} presents a similar analysis using the sPlotOpen dataset, illustrating changes in model response for different plant species when conditioned on the presence of the white oak (\textit{Quercus alba}). Specifically, we highlight plant species for which conditioning on white oak presence results in the largest change in predicted encounter rate compared to the unconditioned case. Notably, the species exhibiting the highest positive changes are consistent with those typically found in oak-hickory forest ecosystems, such as the flowering dogwood and the pignut hickory \citep{fralish12004keystone}.

\begin{figure}
    \centering
    \begin{subfigure}{\textwidth}
        \centering
        \includegraphics[width=\linewidth]{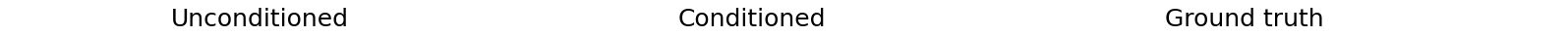}
    \end{subfigure}
    \begin{subfigure}{\textwidth}
        \centering
        \includegraphics[width=\linewidth]{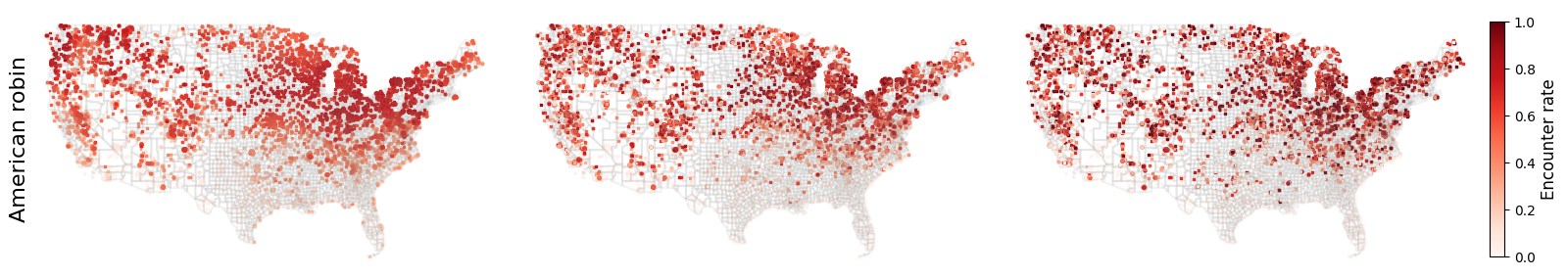}
    \end{subfigure}
    \begin{subfigure}{\textwidth}
        \centering
        \includegraphics[width=\linewidth]{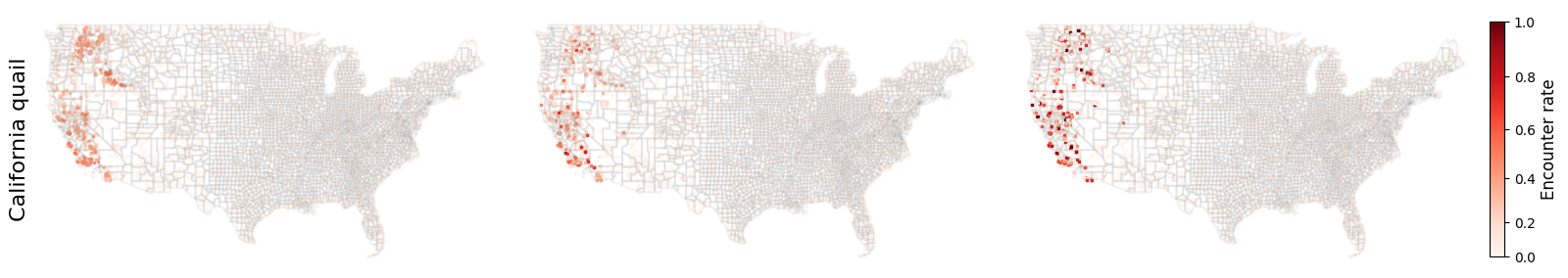}
    \end{subfigure}
    \begin{subfigure}{\textwidth}
        \centering
        \includegraphics[width=\linewidth]{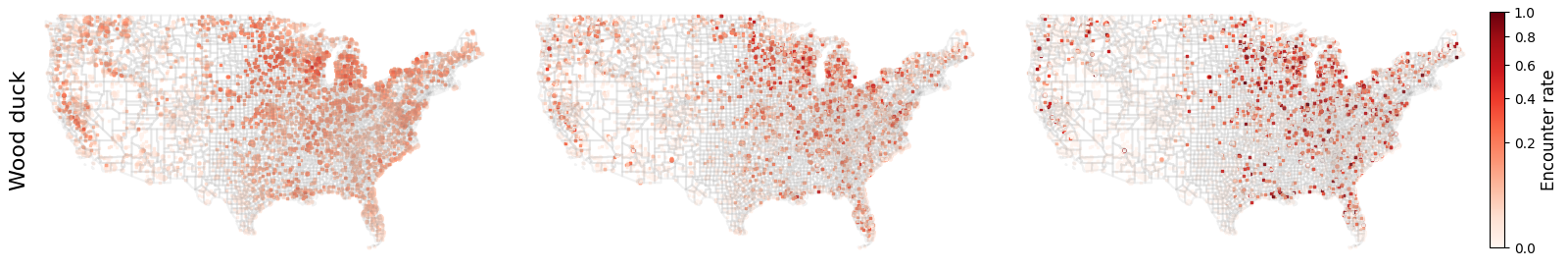}
    \end{subfigure}
    \begin{subfigure}{\textwidth}
        \centering
        \includegraphics[width=\linewidth]{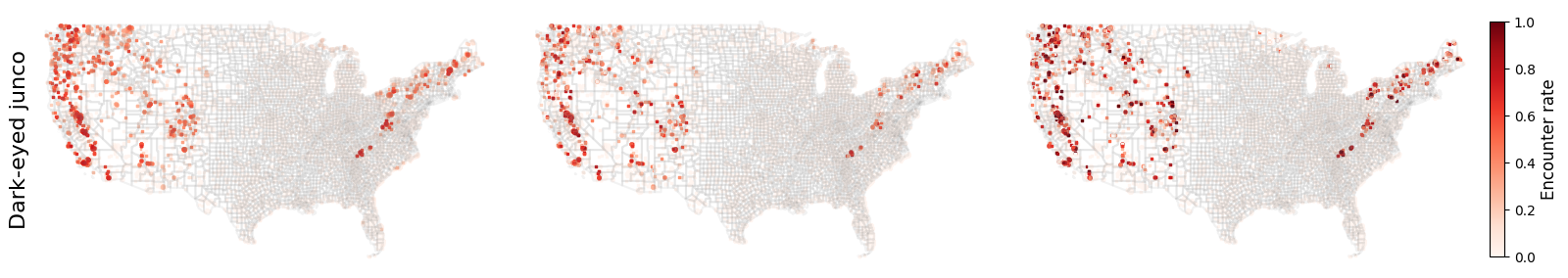}
    \end{subfigure}
    \caption{Encounter rates of test set hotspots: predictions in the unconditioned case (left), predictions when all other bird species' encounter rates are known (middle), and ground truth observations (right).}
    \label{fig:prediction_maps}
\end{figure}

Figure \ref{fig:prediction_maps} shows how encounter rate maps change for a given species when all the other species' encounter rates of the SatBird dataset are known, compared to the unconditioned case, highlighting the impact of biotic variables. The selected species represent different distributional patterns: a widespread species (American robin), a species with a narrower range (California quail), and two species occupying different habitats (the wood duck, typically found in wetlands, and the dark-eyed junco, which breeds in coniferous or mixed-coniferous forests). We observe that the model generally underestimates encounter rates in areas where ground truth encounter rates are higher, both in the unconditioned and conditioned cases. However, conditioning on other species improves predictions, bringing estimated encounter rates closer to ground truth values. This improvement is particularly evident in the more accurate identification of absences, as well as in high encounter rate regions, such as the northeastern US for the American robin and the Sierra Nevada mountains for the dark-eyed junco.

\section{Discussion}
\label{sec:discussion}

In this paper, we propose a novel deep learning-based method, CISO, designed to leverage partial species observations to improve SDMs. CISO conditions its predictions on an arbitrary amount of available information about the presence or absence of other species. This is highly valuable as species observations are often sparse and inconsistent across locations. CISO also enables the modeling of complex, asymmetric relationships between multiple species, capturing the intricacies of ecological interactions. It combines this partial information non-linearly with the abiotic environmental variables. This integration is particularly important, as the strengths of biotic interactions are known to be influenced by environmental conditions \citep{anderson2017and, tikhonov2017using, dormann2018biotic}. Importantly, by deep learning standards, CISO is a lightweight model, and it can be trained on a single consumer-grade GPU, or even on a CPU, albeit with longer training time. 

We explored two distinct scenarios. In the first setup (within-dataset), we predicted species suitability based on the presence or absence of other species within a single dataset, collected using a consistent observational protocol. In the second, more ambitious, across-datasets setup, we combined datasets representing distinct groups of species, where predictions are conditioned on observations from the other group.

Our results demonstrate that incorporating additional information about the presence or absence of other species consistently improves predictive performance in the within-dataset setting, in line with \cite{wisz2013role}, \cite{poggiato2021interpretations}, and \cite{valle2024species} on the role of biotic variables in shaping species distributions. We showed that CISO is an effective method for integrating these variable biotic factors, outperforming alternative approaches. While we evaluated models on scenarios where the sets of known and unknown species labels are fixed at inference time, it is important to note that CISO is inherently flexible. It can accommodate any arbitrary subset of known species occurrences at a given location, including the empty set, making it broadly applicable.

The across-datasets setup poses greater challenges due to the limited number of co-located observations. Unlike the within-dataset setup, where data for all species is available at each site, only a small fraction of locations in the across-datasets scenario contain data from both groups. Nevertheless, in the setup combining the SatBird and SatButterfly datasets, CISO conditioned on the available observations achieves the best performance, on the task of predicting bird species encounter rates from butterfly observations and \textit{vice versa}. This showcases the potential of CISO to condition its predictions on partial observations even across disparate datasets. Notably, SatBird is more than nine times larger than SatButterfly, and co-locations represent only 5\% of the total number of locations covered by both datasets. 

However, when combining the sPlotOpen and SatBird datasets, conditioning on available species degraded performance. While CISO outperformed all baselines in the unconditioned setting, it failed to generalize effectively when conditioned. One reason for this could be the extremely low number of co-locations between the two datasets. Indeed, only 0.5\% of the locations (\num{1161} sites) considered have co-located data, limiting the model's ability to learn meaningful co-occurrence patterns.
Another point to consider is the spatial radius chosen to define co-locations. We did not tune this parameter; yet, the inherent uncertainty in location reporting, along with the possibility that species interactions vary with spatial scales, suggests that further exploration into the definition of co-locations could improve across-datasets modeling.

Besides, since the datasets were collected at different times and through different methodologies—meaning they do not represent actual co-occurrences—we made certain assumptions to enable their integration. Specifically, we assumed that if the temporal gap between observations was not too large and niche equilibrium was maintained (i.e., ecological niches remained relatively stable over time), the datasets could be treated as providing pseudo-co-occurrence data suitable for modeling. However, such integration should be carried out carefully. The validity of the assumptions should always be assessed based on ecological principles and rigorously evaluated with independent data.

Nevertheless, the differential response of predicted species presence to observations of other species could help in the automated identification of potential biotic interactions, providing a starting point for further exploration. It is important to emphasize that co-occurrence patterns alone do not necessarily imply direct ecological interactions \citep{blanchet2020co}. Such patterns can also result from missing biotic or abiotic variables correlated with the conditioning species' presence or absence \citep{dormann2018biotic, valle2024species}. Since our model learns from co-occurrence data rather than causal information, any attempt to infer ecological relationships from its outputs should be approached with caution.

As mentioned earlier, an important limitation of this work is the limited availability of co-located datasets for training and evaluating the model. All approaches to modeling co-occurrences require a large sample size, and CISO is no exception. Nevertheless, the growing volume of species observations collected through citizen science offers promise for improving the modeling of species interactions. An important area for future work is to address the question of imbalance in the datasets to be combined. Datasets often vary in size, and without proper handling of asymmetries in size, the model may neglect taxa with fewer observations.

An even greater challenge is extending our approach to opportunistic presence-only data, which constitutes the majority of citizen science observations, such as those from the iNaturalist platform. The main issue with presence-only data arises when observations are made for a single species, without any co-occurrence information. CISO is designed to learn from the co-occurrence of multiple species observed at the same location and time, rather than from isolated individual observations typical in many presence-only datasets. Therefore, a promising avenue for future exploration is to address the challenge of aggregating such isolated observations and determining under which conditions they can be treated as co-occurrences. Additionally, while the SatBird and SatButterfly datasets used in this study are derived from ``complete checklist'' records from eBird and eButterfly, CISO is also capable of leveraging incomplete checklist data, which are common on these platforms.

We also see great potential for improvement by integrating species characteristics, such as traits, taxonomy, or habitat and range description, into the model through the species embeddings. Such information has been integrated successfully into SDMs in previous studies \citep{pollock2012role, pollock2020protecting, bourhis2023explainable, hamilton2024combining}, and provides valuable context for the model to better understand relationships between species. Finally, beyond individual species characteristics, integrating known species interactions, such as those captured in food webs, has been shown to improve the estimation of realized niches \citep{pellissier2013combining, poggiato2025integrating}. Future work on integrating species interactions into CISO could provide a stronger prior on species relationships, ultimately enhancing model predictions.

\vspace{2em}

\subsection*{Acknowledgments}

This work was partly supported by the Swiss National Science Foundation, under grant 200021\_204057 ``Learning unbiased habitat suitability at scale with AI (deepHSM)'', the Google-Mila partnership program, the Canada CIFAR AI Chairs program, and the Global Center on AI and Biodiversity Change (US NSF OISE-2330423 and Canada NSERC 585136). We also acknowledge computational support from Mila – Quebec AI Institute, including in-kind support from Nvidia Corporation. We thank Catherine Villeneuve and Maxim Larrivée for insightful discussions and the Mila IDT team for their technical support.

\subsection*{Author contributions}

We adhere to the CRediT taxonomy. \textbf{Hager Radi Abdelwahed:} Conceptualization, Data curation, Formal analysis, Investigation, Methodology, Software, Visualization, Writing – original draft, Writing – review \& editing. \textbf{Mélisande Teng:} Conceptualization, Data curation, Formal analysis, Investigation, Methodology, Software, Visualization, Writing – original draft, Writing – review \& editing, \textbf{Robin Zbinden:} Conceptualization, Data curation, Formal analysis, Investigation, Methodology, Software, Visualization, Writing – original draft, Writing – review \& editing. \textbf{Laura Pollock:} Conceptualization, Writing – review \& editing. \textbf{Hugo Larochelle:} Conceptualization, Methodology, Resources, Supervision, Visualization, Writing – review \& editing. \textbf{Devis Tuia:} Conceptualization, Methodology,  Funding acquisition, Resources, Supervision, Writing – review \& editing. \textbf{David Rolnick:} Conceptualization, Methodology, Funding acquisition, Resources, Supervision, Writing – review \& editing.

\subsection*{Data availability statement}

The original SatBird dataset \citep{teng2023satbird} is available at: \url{https://drive.google.com/drive/folders/1eaL2T7U9Imq_CTDSSillETSDJ1vxi5Wq}. We release the SatButterfly dataset at: {\url{https://huggingface.co/datasets/cisosdm/SatButterfly}}.
The sPlotOpen data \citep{sabatini2021splotopen} were downloaded from: \url{https://doi.org/10.25829/idiv.3474-40-3292}.
Environmental variables associated with sPlotOpen observations, including WorldClim v2.1 \citep{fick_worldclim_2017} and SoilGrids data \citep{hengl2017soilgrids250m}, were obtained from: \url{https://www.worldclim.org/data/worldclim21.html} and \url{https://soilgrids.org/}, respectively.
To facilitate the reproduction of experimental results, we provide processed versions of these datasets at: {\url{https://huggingface.co/cisosdm/datasets}} with the code at: {\url{https://anonymous.4open.science/r/SDMPartialLabels-B2EC/}}. All trained model checkpoints are released at: {\url{https://huggingface.co/cisosdm/model_checkpoints}}.


\newpage
\bibliographystyle{abbrvnat}
\bibliography{references}

\begin{thebibliography}{84}
\providecommand{\natexlab}[1]{#1}
\providecommand{\url}[1]{\texttt{#1}}
\expandafter\ifx\csname urlstyle\endcsname\relax
  \providecommand{\doi}[1]{doi: #1}\else
  \providecommand{\doi}{doi: \begingroup \urlstyle{rm}\Url}\fi

\bibitem[Anderson(2017)]{anderson2017and}
R.~P. Anderson.
\newblock When and how should biotic interactions be considered in models of species niches and distributions?
\newblock \emph{Journal of Biogeography}, 44\penalty0 (1):\penalty0 8--17, 2017.

\bibitem[Ara{\'u}jo and Luoto(2007)]{araujo2007importance}
M.~B. Ara{\'u}jo and M.~Luoto.
\newblock The importance of biotic interactions for modelling species distributions under climate change.
\newblock \emph{Global Ecology and Biogeography}, 16\penalty0 (6):\penalty0 743--753, 2007.

\bibitem[Austin and Van~Niel(2011)]{austin2011improving}
M.~P. Austin and K.~P. Van~Niel.
\newblock Improving species distribution models for climate change studies: variable selection and scale, 2011.

\bibitem[Barbet-Massin et~al.(2018)Barbet-Massin, Rome, Villemant, and Courchamp]{barbet2018can}
M.~Barbet-Massin, Q.~Rome, C.~Villemant, and F.~Courchamp.
\newblock Can species distribution models really predict the expansion of invasive species?
\newblock \emph{PloS one}, 13\penalty0 (3):\penalty0 e0193085, 2018.

\bibitem[Blanchet et~al.(2020)Blanchet, Cazelles, and Gravel]{blanchet2020co}
F.~G. Blanchet, K.~Cazelles, and D.~Gravel.
\newblock Co-occurrence is not evidence of ecological interactions.
\newblock \emph{Ecology Letters}, 23\penalty0 (7):\penalty0 1050--1063, 2020.

\bibitem[Borowiec et~al.(2022)Borowiec, Dikow, Frandsen, McKeeken, Valentini, and White]{borowiec2022deep}
M.~L. Borowiec, R.~B. Dikow, P.~B. Frandsen, A.~McKeeken, G.~Valentini, and A.~E. White.
\newblock Deep learning as a tool for ecology and evolution.
\newblock \emph{Methods in Ecology and Evolution}, 13\penalty0 (8):\penalty0 1640--1660, 2022.

\bibitem[Botella et~al.(2018)Botella, Joly, Bonnet, Monestiez, and Munoz]{botella2018deep}
C.~Botella, A.~Joly, P.~Bonnet, P.~Monestiez, and F.~Munoz.
\newblock A deep learning approach to species distribution modelling.
\newblock \emph{Multimedia Tools and Applications for Environmental \& Biodiversity Informatics}, pages 169--199, 2018.

\bibitem[Bourhis et~al.(2023)Bourhis, Bell, Shortall, Kunin, and Milne]{bourhis2023explainable}
Y.~Bourhis, J.~R. Bell, C.~R. Shortall, W.~E. Kunin, and A.~E. Milne.
\newblock Explainable neural networks for trait-based multispecies distribution modelling—a case study with butterflies and moths.
\newblock \emph{Methods in ecology and evolution}, 14\penalty0 (6):\penalty0 1531--1542, 2023.

\bibitem[Bruelheide et~al.(2019)Bruelheide, Dengler, Jim{\'e}nez-Alfaro, Purschke, Hennekens, Chytr{\`y}, Pillar, Jansen, Kattge, Sandel, et~al.]{bruelheide2019splot}
H.~Bruelheide, J.~Dengler, B.~Jim{\'e}nez-Alfaro, O.~Purschke, S.~M. Hennekens, M.~Chytr{\`y}, V.~D. Pillar, F.~Jansen, J.~Kattge, B.~Sandel, et~al.
\newblock splot--a new tool for global vegetation analyses.
\newblock \emph{Journal of Vegetation Science}, 30\penalty0 (2):\penalty0 161--186, 2019.

\bibitem[Brun et~al.(2024)Brun, Karger, Zurell, Descombes, de~Witte, de~Lutio, Wegner, and Zimmermann]{brun2024multispecies}
P.~Brun, D.~N. Karger, D.~Zurell, P.~Descombes, L.~C. de~Witte, R.~de~Lutio, J.~D. Wegner, and N.~E. Zimmermann.
\newblock Multispecies deep learning using citizen science data produces more informative plant community models.
\newblock \emph{Nature Communications}, 15\penalty0 (1):\penalty0 4421, 2024.

\bibitem[Cazzolla~Gatti et~al.(2022)Cazzolla~Gatti, Reich, Gamarra, Crowther, Hui, Morera, Bastin, De-Miguel, Nabuurs, Svenning, et~al.]{cazzolla2022number}
R.~Cazzolla~Gatti, P.~B. Reich, J.~G. Gamarra, T.~Crowther, C.~Hui, A.~Morera, J.-F. Bastin, S.~De-Miguel, G.-J. Nabuurs, J.-C. Svenning, et~al.
\newblock The number of tree species on earth.
\newblock \emph{Proceedings of the National Academy of Sciences}, 119\penalty0 (6):\penalty0 e2115329119, 2022.

\bibitem[Chowdhury et~al.(2021)Chowdhury, Fuller, Dingle, Chapman, and Zalucki]{chowdhury2021migration}
S.~Chowdhury, R.~A. Fuller, H.~Dingle, J.~W. Chapman, and M.~P. Zalucki.
\newblock Migration in butterflies: a global overview.
\newblock \emph{Biological Reviews}, 96\penalty0 (4):\penalty0 1462--1483, 2021.

\bibitem[Cockle et~al.(2011)Cockle, Martin, and Weso{\l}owski]{cockle2011woodpeckers}
K.~L. Cockle, K.~Martin, and T.~Weso{\l}owski.
\newblock Woodpeckers, decay, and the future of cavity-nesting vertebrate communities worldwide.
\newblock \emph{Frontiers in Ecology and the Environment}, 9\penalty0 (7):\penalty0 377--382, 2011.

\bibitem[Cole et~al.(2020)Cole, Deneu, Lorieul, Servajean, Botella, Morris, Jojic, Bonnet, and Joly]{cole2020geolifedata}
E.~Cole, B.~Deneu, T.~Lorieul, M.~Servajean, C.~Botella, D.~Morris, N.~Jojic, P.~Bonnet, and A.~Joly.
\newblock The {GeoLifeCLEF} 2020 dataset.
\newblock \emph{Preprint arXiv:2004.04192}, 2020.

\bibitem[Cole et~al.(2023)Cole, Van~Horn, Lange, Shepard, Leary, Perona, Loarie, and Mac~Aodha]{cole2023spatial}
E.~Cole, G.~Van~Horn, C.~Lange, A.~Shepard, P.~Leary, P.~Perona, S.~Loarie, and O.~Mac~Aodha.
\newblock Spatial implicit neural representations for global-scale species mapping.
\newblock In \emph{International Conference on Machine Learning}, pages 6320--6342. PMLR, 2023.

\bibitem[Davis et~al.(2023)Davis, Bai, Chen, Robinson, Ruiz-Gutierrez, Gomes, and Fink]{davis2023deep}
C.~L. Davis, Y.~Bai, D.~Chen, O.~Robinson, V.~Ruiz-Gutierrez, C.~P. Gomes, and D.~Fink.
\newblock Deep learning with citizen science data enables estimation of species diversity and composition at continental extents.
\newblock \emph{Ecology}, 104\penalty0 (12):\penalty0 e4175, 2023.

\bibitem[Devlin(2018)]{devlin2018bert}
J.~Devlin.
\newblock Bert: Pre-training of deep bidirectional transformers for language understanding.
\newblock \emph{arXiv preprint arXiv:1810.04805}, 2018.

\bibitem[Dhondt(2012)]{dhondt2012interspecific}
A.~A. Dhondt.
\newblock \emph{Interspecific competition in birds}, volume~2.
\newblock Oxford University Press, 2012.

\bibitem[Dormann et~al.(2012)Dormann, Schymanski, Cabral, Chuine, Graham, Hartig, Kearney, Morin, R{\"o}mermann, Schr{\"o}der, et~al.]{dormann2012correlation}
C.~F. Dormann, S.~J. Schymanski, J.~Cabral, I.~Chuine, C.~Graham, F.~Hartig, M.~Kearney, X.~Morin, C.~R{\"o}mermann, B.~Schr{\"o}der, et~al.
\newblock Correlation and process in species distribution models: bridging a dichotomy.
\newblock \emph{Journal of Biogeography}, 39\penalty0 (12):\penalty0 2119--2131, 2012.

\bibitem[Dormann et~al.(2018)Dormann, Bobrowski, Dehling, Harris, Hartig, Lischke, Moretti, Pagel, Pinkert, Schleuning, et~al.]{dormann2018biotic}
C.~F. Dormann, M.~Bobrowski, D.~M. Dehling, D.~J. Harris, F.~Hartig, H.~Lischke, M.~D. Moretti, J.~Pagel, S.~Pinkert, M.~Schleuning, et~al.
\newblock Biotic interactions in species distribution modelling: 10 questions to guide interpretation and avoid false conclusions.
\newblock \emph{Global ecology and biogeography}, 27\penalty0 (9):\penalty0 1004--1016, 2018.

\bibitem[Dosovitskiy(2020)]{dosovitskiy2020image}
A.~Dosovitskiy.
\newblock An image is worth 16x16 words: Transformers for image recognition at scale.
\newblock \emph{arXiv preprint arXiv:2010.11929}, 2020.

\bibitem[Elith and Leathwick(2009)]{elith2009species}
J.~Elith and J.~R. Leathwick.
\newblock Species distribution models: ecological explanation and prediction across space and time.
\newblock \emph{Annual review of ecology, evolution, and systematics}, 40\penalty0 (1):\penalty0 677--697, 2009.

\bibitem[Escamilla~Molgora et~al.(2022)Escamilla~Molgora, Sedda, Diggle, and Atkinson]{escamilla2022taxonomic}
J.~M. Escamilla~Molgora, L.~Sedda, P.~J. Diggle, and P.~M. Atkinson.
\newblock A taxonomic-based joint species distribution model for presence-only data.
\newblock \emph{Journal of the Royal Society Interface}, 19\penalty0 (187):\penalty0 20210681, 2022.

\bibitem[Ester et~al.(1996)Ester, Kriegel, Sander, and Xu]{dbscan}
M.~Ester, H.-P. Kriegel, J.~Sander, and X.~Xu.
\newblock A density-based algorithm for discovering clusters in large spatial databases with noise.
\newblock In \emph{Proceedings of the Second International Conference on Knowledge Discovery and Data Mining}, KDD'96, page 226–231. AAAI Press, 1996.

\bibitem[Fazzari et~al.(2024)Fazzari, Romano, Falchi, and Stefanini]{fazzari2024animal}
E.~Fazzari, D.~Romano, F.~Falchi, and C.~Stefanini.
\newblock Animal behavior analysis methods using deep learning: A survey.
\newblock \emph{arXiv preprint arXiv:2405.14002}, 2024.

\bibitem[Fick and Hijmans(2017)]{fick_worldclim_2017}
S.~E. Fick and R.~J. Hijmans.
\newblock {WorldClim} 2: new 1-km spatial resolution climate surfaces for global land areas.
\newblock \emph{International Journal of Climatology}, 37\penalty0 (12):\penalty0 4302--4315, 2017.
\newblock ISSN 1097-0088.
\newblock \doi{10.1002/joc.5086}.
\newblock URL \url{https://onlinelibrary.wiley.com/doi/abs/10.1002/joc.5086}.
\newblock \_eprint: https://onlinelibrary.wiley.com/doi/pdf/10.1002/joc.5086.

\bibitem[Fourcade et~al.(2018)Fourcade, Besnard, and Secondi]{fourcade2018paintings}
Y.~Fourcade, A.~G. Besnard, and J.~Secondi.
\newblock Paintings predict the distribution of species, or the challenge of selecting environmental predictors and evaluation statistics.
\newblock \emph{Global Ecology and Biogeography}, 27\penalty0 (2):\penalty0 245--256, 2018.

\bibitem[Fralish(2004)]{fralish12004keystone}
J.~S. Fralish.
\newblock The keystone role of oak and hickory in the central hardwood forest.
\newblock In \emph{Upland Oak Ecology Symposium: History, Current Conditions, and Sustainability: Fayetteville, Arkansas, October 7-10, 2002}, number~73 in General Technical Report SRS, page~78. Southern Research Station, 2004.

\bibitem[G{\'o}mez-Aparicio(2009)]{gomez2009role}
L.~G{\'o}mez-Aparicio.
\newblock The role of plant interactions in the restoration of degraded ecosystems: a meta-analysis across life-forms and ecosystems.
\newblock \emph{Journal of Ecology}, 97\penalty0 (6):\penalty0 1202--1214, 2009.

\bibitem[Gorishniy et~al.(2021)Gorishniy, Rubachev, Khrulkov, and Babenko]{gorishniy2021revisiting}
Y.~Gorishniy, I.~Rubachev, V.~Khrulkov, and A.~Babenko.
\newblock Revisiting deep learning models for tabular data.
\newblock \emph{Advances in Neural Information Processing Systems}, 34:\penalty0 18932--18943, 2021.

\bibitem[Gorishniy et~al.(2022)Gorishniy, Rubachev, and Babenko]{gorishniy2022embeddings}
Y.~Gorishniy, I.~Rubachev, and A.~Babenko.
\newblock On embeddings for numerical features in tabular deep learning.
\newblock \emph{Advances in Neural Information Processing Systems}, 35:\penalty0 24991--25004, 2022.

\bibitem[Guisan and Thuiller(2005)]{guisan2005predicting}
A.~Guisan and W.~Thuiller.
\newblock Predicting species distribution: offering more than simple habitat models.
\newblock \emph{Ecology letters}, 8\penalty0 (9):\penalty0 993--1009, 2005.

\bibitem[Guisan et~al.(2013)Guisan, Tingley, Baumgartner, Naujokaitis-Lewis, Sutcliffe, Tulloch, Regan, Brotons, McDonald-Madden, Mantyka-Pringle, et~al.]{guisan2013predicting}
A.~Guisan, R.~Tingley, J.~B. Baumgartner, I.~Naujokaitis-Lewis, P.~R. Sutcliffe, A.~I. Tulloch, T.~J. Regan, L.~Brotons, E.~McDonald-Madden, C.~Mantyka-Pringle, et~al.
\newblock Predicting species distributions for conservation decisions.
\newblock \emph{Ecology letters}, 16\penalty0 (12):\penalty0 1424--1435, 2013.

\bibitem[Hamilton et~al.(2024)Hamilton, Lange, Cole, Shepard, Heinrich, Mac~Aodha, Van~Horn, and Maji]{hamilton2024combining}
M.~Hamilton, C.~Lange, E.~Cole, A.~Shepard, S.~Heinrich, O.~Mac~Aodha, G.~Van~Horn, and S.~Maji.
\newblock Combining observational data and language for species range estimation.
\newblock \emph{arXiv preprint arXiv:2410.10931}, 2024.

\bibitem[Harris(2015)]{harris2015generating}
D.~J. Harris.
\newblock Generating realistic assemblages with a joint species distribution model.
\newblock \emph{Methods in Ecology and Evolution}, 6\penalty0 (4):\penalty0 465--473, 2015.

\bibitem[Hengl et~al.(2017)Hengl, Mendes~de Jesus, Heuvelink, Ruiperez~Gonzalez, Kilibarda, Blagoti{\'c}, Shangguan, Wright, Geng, Bauer-Marschallinger, et~al.]{hengl2017soilgrids250m}
T.~Hengl, J.~Mendes~de Jesus, G.~B. Heuvelink, M.~Ruiperez~Gonzalez, M.~Kilibarda, A.~Blagoti{\'c}, W.~Shangguan, M.~N. Wright, X.~Geng, B.~Bauer-Marschallinger, et~al.
\newblock Soilgrids250m: Global gridded soil information based on machine learning.
\newblock \emph{PLoS one}, 12\penalty0 (2):\penalty0 e0169748, 2017.

\bibitem[Hijmans and al.(2005)]{oldworldclim}
Hijmans and al.
\newblock Worldclim 1.4 (historical climate conditions).
\newblock \emph{International journal of climatology}, 25, 2005.

\bibitem[Hortal et~al.(2015)Hortal, de~Bello, Diniz-Filho, Lewinsohn, Lobo, and Ladle]{hortal2015seven}
J.~Hortal, F.~de~Bello, J.~A.~F. Diniz-Filho, T.~M. Lewinsohn, J.~M. Lobo, and R.~J. Ladle.
\newblock Seven shortfalls that beset large-scale knowledge of biodiversity.
\newblock \emph{Annual review of ecology, evolution, and systematics}, 46\penalty0 (1):\penalty0 523--549, 2015.

\bibitem[Jetz et~al.(2019)Jetz, McGeoch, Guralnick, Ferrier, Beck, Costello, Fernandez, Geller, Keil, Merow, et~al.]{jetz2019essential}
W.~Jetz, M.~A. McGeoch, R.~Guralnick, S.~Ferrier, J.~Beck, M.~J. Costello, M.~Fernandez, G.~N. Geller, P.~Keil, C.~Merow, et~al.
\newblock Essential biodiversity variables for mapping and monitoring species populations.
\newblock \emph{Nature ecology \& evolution}, 3\penalty0 (4):\penalty0 539--551, 2019.

\bibitem[Johnston et~al.(2021)Johnston, Hochachka, Strimas-Mackey, Ruiz~Gutierrez, Robinson, Miller, Auer, Kelling, and Fink]{johnston2021analytical}
A.~Johnston, W.~M. Hochachka, M.~E. Strimas-Mackey, V.~Ruiz~Gutierrez, O.~J. Robinson, E.~T. Miller, T.~Auer, S.~T. Kelling, and D.~Fink.
\newblock Analytical guidelines to increase the value of community science data: An example using ebird data to estimate species distributions.
\newblock \emph{Diversity and Distributions}, 27\penalty0 (7):\penalty0 1265--1277, 2021.

\bibitem[Joseph(2020)]{joseph2020neural}
M.~B. Joseph.
\newblock Neural hierarchical models of ecological populations.
\newblock \emph{Ecology Letters}, 23\penalty0 (4):\penalty0 734--747, 2020.

\bibitem[Kahl et~al.(2021)Kahl, Wood, Eibl, and Klinck]{kahl2021birdnet}
S.~Kahl, C.~M. Wood, M.~Eibl, and H.~Klinck.
\newblock Birdnet: A deep learning solution for avian diversity monitoring.
\newblock \emph{Ecological Informatics}, 61:\penalty0 101236, 2021.

\bibitem[Kellenberger et~al.(2024)Kellenberger, Winner, and Jetz]{kellenberger2024performance}
B.~A. Kellenberger, K.~Winner, and W.~Jetz.
\newblock The performance and potential of deep learning for predicting species distributions.
\newblock \emph{bioRxiv}, pages 2024--08, 2024.

\bibitem[Kelling et~al.(2013)Kelling, Gerbracht, Fink, Lagoze, Wong, Yu, Damoulas, and Gomes]{eBird:HCLN}
S.~Kelling, J.~Gerbracht, D.~Fink, C.~Lagoze, W.-K. Wong, J.~Yu, T.~Damoulas, and C.~Gomes.
\newblock {eBird}: A human/computer learning network for biodiversity conservation and research.
\newblock \emph{AI Magazine}, 34, 03 2013.

\bibitem[Lanchantin et~al.(2021)Lanchantin, Wang, Ordonez, and Qi]{cTran}
J.~Lanchantin, T.~Wang, V.~Ordonez, and Y.~Qi.
\newblock General multi-label image classification with transformers.
\newblock In \emph{Proceedings of the IEEE/CVF Conference on Computer Vision and Pattern Recognition (CVPR)}, pages 16478--16488, June 2021.

\bibitem[Lange et~al.(2024)Lange, Cole, Horn, and Mac~Aodha]{lange2024active}
C.~Lange, E.~Cole, G.~Horn, and O.~Mac~Aodha.
\newblock Active learning-based species range estimation.
\newblock \emph{Advances in Neural Information Processing Systems}, 36, 2024.

\bibitem[Loshchilov and Hutter(2019)]{adamW}
I.~Loshchilov and F.~Hutter.
\newblock Decoupled weight decay regularization.
\newblock In \emph{International Conference on Learning Representations}, 2019.
\newblock URL \url{https://openreview.net/forum?id=Bkg6RiCqY7}.

\bibitem[Mod et~al.(2016)Mod, Scherrer, Luoto, and Guisan]{mod2016we}
H.~K. Mod, D.~Scherrer, M.~Luoto, and A.~Guisan.
\newblock What we use is not what we know: environmental predictors in plant distribution models.
\newblock \emph{Journal of Vegetation Science}, 27\penalty0 (6):\penalty0 1308--1322, 2016.

\bibitem[Norouzzadeh et~al.(2021)Norouzzadeh, Morris, Beery, Joshi, Jojic, and Clune]{norouzzadeh2021deep}
M.~S. Norouzzadeh, D.~Morris, S.~Beery, N.~Joshi, N.~Jojic, and J.~Clune.
\newblock A deep active learning system for species identification and counting in camera trap images.
\newblock \emph{Methods in ecology and evolution}, 12\penalty0 (1):\penalty0 150--161, 2021.

\bibitem[Omohundro(1989)]{omohundro1989five}
S.~M. Omohundro.
\newblock Five balltree construction algorithms, 1989.

\bibitem[Ovaskainen et~al.(2016)Ovaskainen, Abrego, Halme, and Dunson]{ovaskainen2016using}
O.~Ovaskainen, N.~Abrego, P.~Halme, and D.~Dunson.
\newblock Using latent variable models to identify large networks of species-to-species associations at different spatial scales.
\newblock \emph{Methods in Ecology and Evolution}, 7\penalty0 (5):\penalty0 549--555, 2016.

\bibitem[Padilla and Pugnaire(2006)]{padilla2006role}
F.~M. Padilla and F.~I. Pugnaire.
\newblock The role of nurse plants in the restoration of degraded environments.
\newblock \emph{Frontiers in Ecology and the Environment}, 4\penalty0 (4):\penalty0 196--202, 2006.

\bibitem[Pearson and Dawson(2003)]{pearson2003predicting}
R.~G. Pearson and T.~P. Dawson.
\newblock Predicting the impacts of climate change on the distribution of species: are bioclimate envelope models useful?
\newblock \emph{Global ecology and biogeography}, 12\penalty0 (5):\penalty0 361--371, 2003.

\bibitem[Pellissier et~al.(2010)Pellissier, Anne~Br{\aa}then, Pottier, Randin, Vittoz, Dubuis, Yoccoz, Alm, Zimmermann, and Guisan]{pellissier2010species}
L.~Pellissier, K.~Anne~Br{\aa}then, J.~Pottier, C.~F. Randin, P.~Vittoz, A.~Dubuis, N.~G. Yoccoz, T.~Alm, N.~E. Zimmermann, and A.~Guisan.
\newblock Species distribution models reveal apparent competitive and facilitative effects of a dominant species on the distribution of tundra plants.
\newblock \emph{Ecography}, 33\penalty0 (6):\penalty0 1004--1014, 2010.

\bibitem[Pellissier et~al.(2013)Pellissier, Rohr, Ndiribe, Pradervand, Salamin, Guisan, and Wisz]{pellissier2013combining}
L.~Pellissier, R.~P. Rohr, C.~Ndiribe, J.-N. Pradervand, N.~Salamin, A.~Guisan, and M.~Wisz.
\newblock Combining food web and species distribution models for improved community projections.
\newblock \emph{Ecology and evolution}, 3\penalty0 (13):\penalty0 4572--4583, 2013.

\bibitem[Phillips et~al.(2006)Phillips, Anderson, and Schapire]{phillips2006maximum}
S.~J. Phillips, R.~P. Anderson, and R.~E. Schapire.
\newblock Maximum entropy modeling of species geographic distributions.
\newblock \emph{Ecological modelling}, 190\penalty0 (3-4):\penalty0 231--259, 2006.

\bibitem[Picek et~al.(2024)Picek, Botella, Servajean, Leblanc, Palard, Larcher, Deneu, Marcos, Bonnet, and Joly]{picek2024geoplant}
L.~Picek, C.~Botella, M.~Servajean, C.~Leblanc, R.~Palard, T.~Larcher, B.~Deneu, D.~Marcos, P.~Bonnet, and A.~Joly.
\newblock Geoplant: Spatial plant species prediction dataset.
\newblock \emph{arXiv preprint arXiv:2408.13928}, 2024.

\bibitem[Poggiato et~al.(2021)Poggiato, M{\"u}nkem{\"u}ller, Bystrova, Arbel, Clark, and Thuiller]{poggiato2021interpretations}
G.~Poggiato, T.~M{\"u}nkem{\"u}ller, D.~Bystrova, J.~Arbel, J.~S. Clark, and W.~Thuiller.
\newblock On the interpretations of joint modeling in community ecology.
\newblock \emph{Trends in ecology \& evolution}, 36\penalty0 (5):\penalty0 391--401, 2021.

\bibitem[Poggiato et~al.(2025)Poggiato, Andr{\'e}oletti, Pollock, and Thuiller]{poggiato2025integrating}
G.~Poggiato, J.~Andr{\'e}oletti, L.~J. Pollock, and W.~Thuiller.
\newblock Integrating food webs in species distribution models can improve ecological niche estimation and predictions.
\newblock \emph{Ecography}, page e07546, 2025.

\bibitem[Pollock et~al.(2012)Pollock, Morris, and Vesk]{pollock2012role}
L.~J. Pollock, W.~K. Morris, and P.~A. Vesk.
\newblock The role of functional traits in species distributions revealed through a hierarchical model.
\newblock \emph{Ecography}, 35\penalty0 (8):\penalty0 716--725, 2012.

\bibitem[Pollock et~al.(2014)Pollock, Tingley, Morris, Golding, O'Hara, Parris, Vesk, and McCarthy]{pollock2014understanding}
L.~J. Pollock, R.~Tingley, W.~K. Morris, N.~Golding, R.~B. O'Hara, K.~M. Parris, P.~A. Vesk, and M.~A. McCarthy.
\newblock Understanding co-occurrence by modelling species simultaneously with a joint species distribution model (jsdm).
\newblock \emph{Methods in Ecology and Evolution}, 5\penalty0 (5):\penalty0 397--406, 2014.

\bibitem[Pollock et~al.(2020)Pollock, O’connor, Mokany, Rosauer, Talluto, and Thuiller]{pollock2020protecting}
L.~J. Pollock, L.~M. O’connor, K.~Mokany, D.~F. Rosauer, M.~V. Talluto, and W.~Thuiller.
\newblock Protecting biodiversity (in all its complexity): new models and methods.
\newblock \emph{Trends in Ecology \& Evolution}, 35\penalty0 (12):\penalty0 1119--1128, 2020.

\bibitem[Pollock et~al.(2025)Pollock, Kitzes, Beery, Gaynor, Jarzyna, Mac~Aodha, Meyer, Rolnick, Taylor, Tuia, et~al.]{pollock2025harnessing}
L.~J. Pollock, J.~Kitzes, S.~Beery, K.~M. Gaynor, M.~A. Jarzyna, O.~Mac~Aodha, B.~Meyer, D.~Rolnick, G.~W. Taylor, D.~Tuia, et~al.
\newblock Harnessing artificial intelligence to fill global shortfalls in biodiversity knowledge.
\newblock \emph{Nature Reviews Biodiversity}, pages 1--17, 2025.

\bibitem[Prudic et~al.(2017)Prudic, McFarland, Oliver, Hutchinson, Long, Kerr, and Larriv{\'e}e]{prudic2017ebutterfly}
K.~L. Prudic, K.~P. McFarland, J.~C. Oliver, R.~A. Hutchinson, E.~C. Long, J.~T. Kerr, and M.~Larriv{\'e}e.
\newblock ebutterfly: leveraging massive online citizen science for butterfly conservation.
\newblock \emph{Insects}, 8\penalty0 (2):\penalty0 53, 2017.

\bibitem[Roberts et~al.(2017)Roberts, Bahn, Ciuti, Boyce, Elith, Guillera-Arroita, Hauenstein, Lahoz-Monfort, Schr{\"o}der, Thuiller, et~al.]{roberts2017cross}
D.~R. Roberts, V.~Bahn, S.~Ciuti, M.~S. Boyce, J.~Elith, G.~Guillera-Arroita, S.~Hauenstein, J.~J. Lahoz-Monfort, B.~Schr{\"o}der, W.~Thuiller, et~al.
\newblock Cross-validation strategies for data with temporal, spatial, hierarchical, or phylogenetic structure.
\newblock \emph{Ecography}, 40\penalty0 (8):\penalty0 913--929, 2017.

\bibitem[Romera-Romera and Nieto-Lugilde(2024)]{romera2024should}
D.~Romera-Romera and D.~Nieto-Lugilde.
\newblock Should we exploit opportunistic databases with joint species distribution models? artificial and real data suggest it depends on the sampling completeness.
\newblock \emph{Ecography}, page e07340, 2024.

\bibitem[Romera-Romera and Nieto-Lugilde(2025)]{romera2025should}
D.~Romera-Romera and D.~Nieto-Lugilde.
\newblock Should we exploit opportunistic databases with joint species distribution models? artificial and real data suggest it depends on the sampling completeness.
\newblock \emph{Ecography}, 2025\penalty0 (2):\penalty0 e07340, 2025.

\bibitem[Sabatini et~al.(2021)Sabatini, Lenoir, Hattab, Arnst, Chytr{\`y}, Dengler, De~Ruffray, Hennekens, Jandt, Jansen, et~al.]{sabatini2021splotopen}
F.~M. Sabatini, J.~Lenoir, T.~Hattab, E.~A. Arnst, M.~Chytr{\`y}, J.~Dengler, P.~De~Ruffray, S.~M. Hennekens, U.~Jandt, F.~Jansen, et~al.
\newblock splotopen--an environmentally balanced, open-access, global dataset of vegetation plots.
\newblock \emph{Global Ecology and Biogeography}, 30\penalty0 (9):\penalty0 1740--1764, 2021.

\bibitem[Teng et~al.(2023)Teng, Elmustafa, Akera, Bengio, Radi, Larochelle, and Rolnick]{teng2023satbird}
M.~Teng, A.~Elmustafa, B.~Akera, Y.~Bengio, H.~Radi, H.~Larochelle, and D.~Rolnick.
\newblock Satbird: a dataset for bird species distribution modeling using remote sensing and citizen science data.
\newblock In \emph{Thirty-seventh Conference on Neural Information Processing Systems Datasets and Benchmarks Track}, 2023.
\newblock URL \url{https://openreview.net/forum?id=Vn5qZGxGj3}.

\bibitem[Tiel et~al.(2025)Tiel, Zbinden, Dalsasso, Kellenberger, Pellissier, and Tuia]{tiel2025multi}
N.~v. Tiel, R.~Zbinden, E.~Dalsasso, B.~Kellenberger, L.~Pellissier, and D.~Tuia.
\newblock Multi-scale and multimodal species distribution modeling.
\newblock In \emph{European Conference on Computer Vision}, pages 151--159. Springer, 2025.

\bibitem[Tikhonov et~al.(2017)Tikhonov, Abrego, Dunson, and Ovaskainen]{tikhonov2017using}
G.~Tikhonov, N.~Abrego, D.~Dunson, and O.~Ovaskainen.
\newblock Using joint species distribution models for evaluating how species-to-species associations depend on the environmental context.
\newblock \emph{Methods in Ecology and Evolution}, 8\penalty0 (4):\penalty0 443--452, 2017.

\bibitem[Tuia et~al.(2022)Tuia, Kellenberger, Beery, Costelloe, Zuffi, Risse, Mathis, Mathis, van Langevelde, Burghardt, et~al.]{tuia2022perspectives}
D.~Tuia, B.~Kellenberger, S.~Beery, B.~R. Costelloe, S.~Zuffi, B.~Risse, A.~Mathis, M.~W. Mathis, F.~van Langevelde, T.~Burghardt, et~al.
\newblock Perspectives in machine learning for wildlife conservation.
\newblock \emph{Nature communications}, 13\penalty0 (1):\penalty0 792, 2022.

\bibitem[{U.S. Census Bureau}(2023)]{Census_Bureau_USA_bounderies_shape_file}
{U.S. Census Bureau}.
\newblock {Cartographic Boundary Files - Shapefile}.
\newblock \url{https://www.census.gov/geographies/mapping-files/time-series/geo/carto-boundary-file.html}, 2023.
\newblock Accessed: 2023-06-06.

\bibitem[Vall{\'e} et~al.(2024)Vall{\'e}, Poggiato, Thuiller, Jiguet, Princ{\'e}, and Le~Viol]{valle2024species}
C.~Vall{\'e}, G.~Poggiato, W.~Thuiller, F.~Jiguet, K.~Princ{\'e}, and I.~Le~Viol.
\newblock Species associations in joint species distribution models: from missing variables to conditional predictions.
\newblock \emph{Journal of Biogeography}, 51\penalty0 (2):\penalty0 311--324, 2024.

\bibitem[Vaswani et~al.(2017)Vaswani, Shazeer, Parmar, Uszkoreit, Jones, Gomez, Kaiser, and Polosukhin]{vaswani2017attention}
A.~Vaswani, N.~Shazeer, N.~Parmar, J.~Uszkoreit, L.~Jones, A.~N. Gomez, {\L}.~Kaiser, and I.~Polosukhin.
\newblock Attention is all you need.
\newblock \emph{Advances in neural information processing systems}, 30, 2017.

\bibitem[Wenny(2001)]{wenny2001advantages}
D.~G. Wenny.
\newblock Advantages of seed dispersal: a re-evaluation of directed dispersal.
\newblock \emph{Evolutionary Ecology Research}, 3\penalty0 (1):\penalty0 37--50, 2001.

\bibitem[Wilkinson et~al.(2021)Wilkinson, Golding, Guillera-Arroita, Tingley, and McCarthy]{wilkinson2021defining}
D.~P. Wilkinson, N.~Golding, G.~Guillera-Arroita, R.~Tingley, and M.~A. McCarthy.
\newblock Defining and evaluating predictions of joint species distribution models.
\newblock \emph{Methods in Ecology and Evolution}, 12\penalty0 (3):\penalty0 394--404, 2021.

\bibitem[Wisz et~al.(2013)Wisz, Pottier, Kissling, Pellissier, Lenoir, Damgaard, Dormann, Forchhammer, Grytnes, Guisan, et~al.]{wisz2013role}
M.~S. Wisz, J.~Pottier, W.~D. Kissling, L.~Pellissier, J.~Lenoir, C.~F. Damgaard, C.~F. Dormann, M.~C. Forchhammer, J.-A. Grytnes, A.~Guisan, et~al.
\newblock The role of biotic interactions in shaping distributions and realised assemblages of species: implications for species distribution modelling.
\newblock \emph{Biological reviews}, 88\penalty0 (1):\penalty0 15--30, 2013.

\bibitem[Wood et~al.(2011)Wood, Pidgeon, Gratton, and Wilder]{wood2011effects}
E.~M. Wood, A.~M. Pidgeon, C.~Gratton, and T.~T. Wilder.
\newblock Effects of oak barrens habitat management for karner blue butterfly (lycaeides samuelis) on the avian community.
\newblock \emph{Biological Conservation}, 144\penalty0 (12):\penalty0 3117--3126, 2011.

\bibitem[Zbinden et~al.(2023)Zbinden, Van~Tiel, Kellenberger, Hughes, and Tuia]{zbinden2023exploring}
R.~Zbinden, N.~M.~A. Van~Tiel, B.~A. Kellenberger, L.~Hughes, and D.~Tuia.
\newblock Exploring neural networks and their potential for species distribution modeling.
\newblock In \emph{11th International Conference on Learning Representations (ICLR) Workshops}, 2023.

\bibitem[Zbinden et~al.(2024)Zbinden, Van~Tiel, Kellenberger, Hughes, and Tuia]{zbinden2024selection}
R.~Zbinden, N.~Van~Tiel, B.~Kellenberger, L.~Hughes, and D.~Tuia.
\newblock On the selection and effectiveness of pseudo-absences for species distribution modeling with deep learning.
\newblock \emph{Ecological Informatics}, 81:\penalty0 102623, 2024.

\bibitem[Zbinden et~al.(2025)Zbinden, van Tiel, Sumbul, Vanalli, Kellenberger, and Tuia]{zbinden2025masksdm}
R.~Zbinden, N.~van Tiel, G.~Sumbul, C.~Vanalli, B.~Kellenberger, and D.~Tuia.
\newblock Masksdm with shapley values to improve flexibility, robustness, and explainability in species distribution modeling.
\newblock \emph{arXiv preprint arXiv:2503.13057}, 2025.

\bibitem[Zurell(2017)]{zurell2017integrating}
D.~Zurell.
\newblock Integrating demography, dispersal and interspecific interactions into bird distribution models.
\newblock \emph{Journal of Avian Biology}, 48\penalty0 (12):\penalty0 1505--1516, 2017.

\bibitem[Zurell et~al.(2018)Zurell, Pollock, and Thuiller]{zurell2018joint}
D.~Zurell, L.~J. Pollock, and W.~Thuiller.
\newblock Do joint species distribution models reliably detect interspecific interactions from co-occurrence data in homogenous environments?
\newblock \emph{Ecography}, 41\penalty0 (11):\penalty0 1812--1819, 2018.

\end{thebibliography}

\newpage
\begin{appendix}
\section{Supplementary information on datasets}

\subsection{sPlotOpen}\label{appendix:splotopen}

\begin{figure}[ht]
    \centering
    \includegraphics[width=1\linewidth]{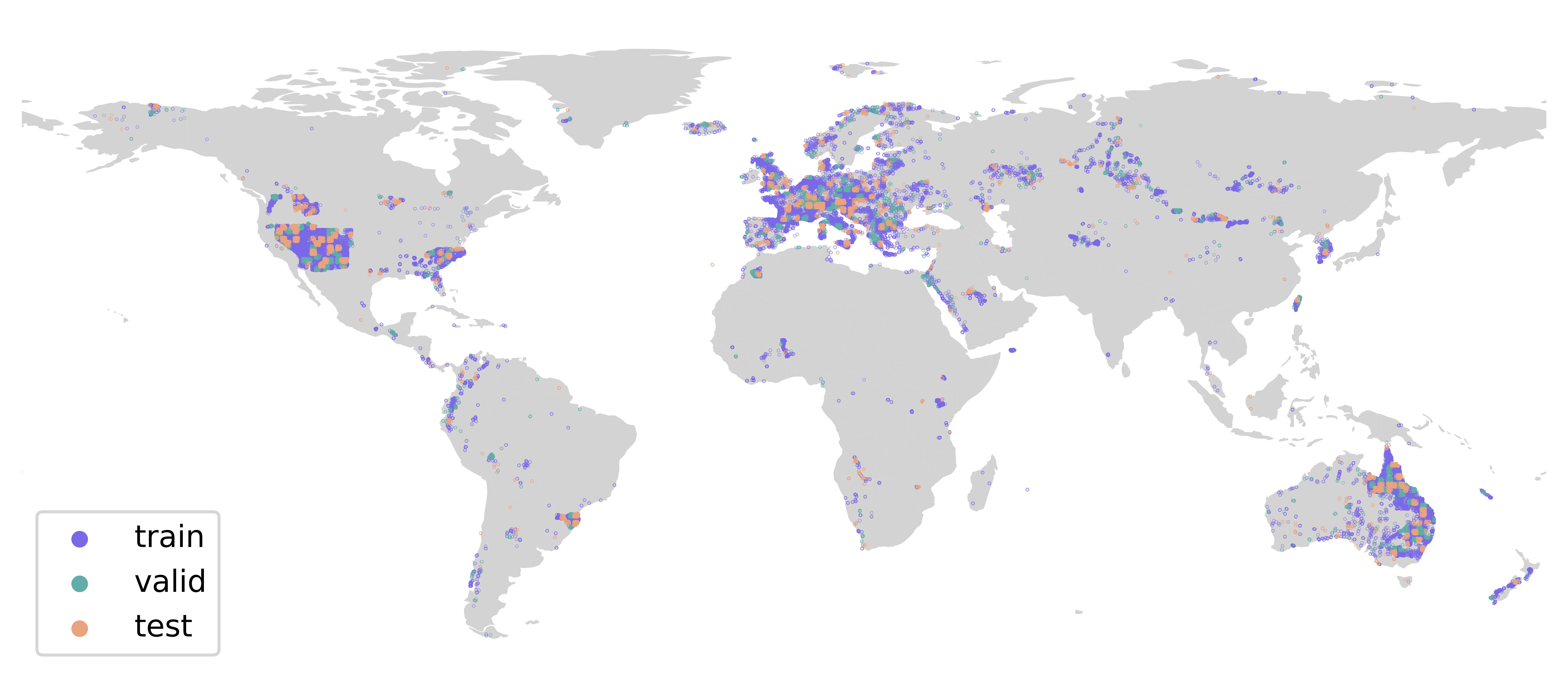}
    \caption{Geographic distribution of training, validation, and test splits in the sPlotOpen dataset.}
    \label{fig:splits_sPlotOpen}
\end{figure}

Figure \ref{fig:splits_sPlotOpen} shows the geographic distribution of the sPlotOpen plot splits obtained using spatial block-cross validation \citep{roberts2017cross}. The splits are distributed as follows 70\% for training, 15\% for validation, and 15\% for testing. Each spatial block covers an area of 1° × 1°.

Table \ref{tab:fuzz_matching} presents the results of species name matching using the \texttt{thefuzz\footnote{\url{https://github.com/seatgeek/thefuzz}}} python package. We include species pairs with a fuzzy score similarity greater 94\%. A total of 11 species pairs were merged, with merging decisions made manually.

\begin{table}
    \renewcommand{\arraystretch}{1.1}
    \centering
    \setlength\tabcolsep{4.8pt}
    \begin{tabular}{l l c c c c c}
        \hline
        species 1 & species 2 & fuzzy score & \#occ & \#occ 1 & \#occ 2 & merged \\
        \hline
        \textit{Echinochloa crus-galli} & \textit{Echinochloa crusgalli} & 98 & 214 & 212 & 2 & yes \\
        \textit{Hypoestes forskaolii} & \textit{Hypoestes forsskaolii} & 98 & 147 & 143 & 4 & yes \\
        \textit{Scutellaria scordifolia} & \textit{Scutellaria scordiifolia} & 98 & 108 & 31 & 77 & yes \\
        \textit{Carex caespitosa} & \textit{Carex cespitosa} & 97 & 165 & 6 & 159 & yes \\
        \textit{Astragalus} & \textit{Astragalus} & 95 & 149 & 7 & 142 & yes \\
        \textit{Calamagrostis epigeios} & \textit{Calamagrostis epigejos} & 95 & 1070 & 584 & 486 & yes \\
        \textit{Solidago virga-aurea} & \textit{Solidago virgaurea} & 95 & 3821 & 83 & 3738 & yes \\
        \textit{Cabralea cangerana} & \textit{Cabralea canjerana} & 94 & 607 & 6 & 601 & yes \\
        \textit{Uraria lagopodioides} & \textit{Uraria lagopodoides} & 97 & 135 & 134 & 1 & yes \\
        \textit{Schinus terebinthifolia} & \textit{Schinus terebinthifolius} & 94 & 153 & 78 & 75 & yes \\
        \textit{Leontodon hispidulus} & \textit{Leontodon hispidus} & 95 & 1574 & 1 & 1573 & yes \\
        \textit{Trifolium pannonicum} & \textit{Tripolium pannonicum} & 95 & 315 & 19 & 296 & no \\
        \textit{Carex acuta} & \textit{Carex acutata} & 92 & 390 & 388 & 2 & no \\
        \textit{Poa alpigena} & \textit{Poa alpina} & 91 & 1912 & 124 & 1788 & no \\
        \textit{Artemisia frigida} & \textit{Artemisia rigida} & 97 & 1259 & 1257 & 2 & no \\
        \textit{Festuca alaica} & \textit{Festuca altaica} & 97 & 337 & 52 & 285 & no \\
        \textit{Potentilla erecta} & \textit{Potentilla recta} & 97 & 4378 & 4241 & 137 & no \\
        \textit{Blackstonia imperfoliata} & \textit{Blackstonia perfoliata} & 96 & 238 & 6 & 232 & no \\
        \textit{Amelanchier obovalis} & \textit{Amelanchier ovalis} & 95 & 477 & 14 & 463 & no \\
        \textit{Festuca brachyphylla} & \textit{Festuca trachyphylla} & 95 & 263 & 212 & 51 & no \\
        \textit{Hibiscus macranthus} & \textit{Hibiscus micranthus} & 95 & 180 & 22 & 158 & no \\
        \textit{Phyllanthus virgatus} & \textit{Phyllanthus virgulatus} & 95 & 1061 & 1055 & 6 & no \\
        \textit{Ranunculus fontanus} & \textit{Ranunculus montanus} & 95 & 777 & 1 & 776 & no \\
        \textit{Alchemilla alpigena} & \textit{Alchemilla alpina} & 94 & 665 & 63 & 602 & no \\
        \textit{Astragalus glycyphylloides} & \textit{Astragalus glycyphyllos} & 94 & 135 & 4 & 131 & no \\
        \textit{Blepharis mitrata} & \textit{Blepharis obmitrata} & 94 & 112 & 8 & 104 & no \\
        \textit{Carex amgunensis} & \textit{Carex argunensis} & 94 & 180 & 11 & 169 & no \\
        \textit{Dysphania aristata} & \textit{Dysphania cristata} & 94 & 368 & 346 & 22 & no \\
        \textit{Euphorbia hirsuta} & \textit{Euphorbia hirta} & 94 & 109 & 13 & 96 & no \\
        \textit{Ononis minutissima} & \textit{Ononis mitissima} & 94 & 176 & 175 & 1 & no \\
        \textit{Rhynchospora heterochaeta} & \textit{Rhynchospora pterochaeta} & 94 & 122 & 71 & 51 & no \\
        \textit{Trifolium patens} & \textit{Trifolium pratense} & 94 & 3125 & 106 & 3019 & no \\
        \textit{Vernicia montana} & \textit{Veronica montana} & 94 & 250 & 16 & 234 & no \\
        \hline
    \end{tabular}
    \caption{Fuzzy matching results showing similarity scores between species, with corresponding numbers of presence records and merging decisions.}
    \label{tab:fuzz_matching}
\end{table}

\subsection{SatButterfly Dataset}\label{appendix_satbutterfly_dataset}

In this section, we provide additional details on the preparation of the SatButterfly dataset. Specifically, we exclude locations over the sea by applying the cartographic boundaries of the United States at a $5$ meter resolution, as provided by the Census Bureau’s MAF/TIGER \citep{Census_Bureau_USA_bounderies_shape_file}. Unlike Satbird, we retained all the locations that contain at least one complete checklist, given the relative scarcity of butterfly observations.

To align with the predictors provided by SatBird, we also collect and release satellite imagery from the Sentinel-2 satellite and patches from environmental raster datasets, alongside the tabular data used in this study. For each location with observations, we provide a patch of values centered on the site, with dimensions of $50\times 50$. These patches have a spatial resolution of approximately \num{1} km for WorldClim 1.4 \citep{oldworldclim} and \num{250} meters for SoilGrids \citep{hengl2017soilgrids250m}.

\begin{figure}
    \centering
    \includegraphics[width=1\linewidth]{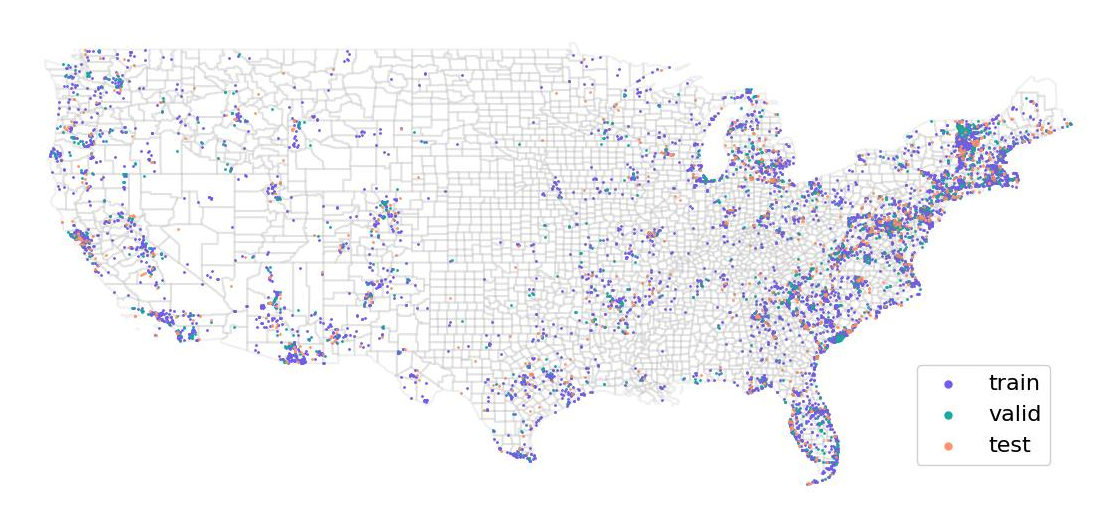}
    \caption{Geographic distribution of training, validation, and test splits in the SatButterfly-v1 dataset.}
    \label{fig:satbutterfly_v1_datadist}
\end{figure}

\begin{figure}
    \centering
    \includegraphics[width=1\linewidth]{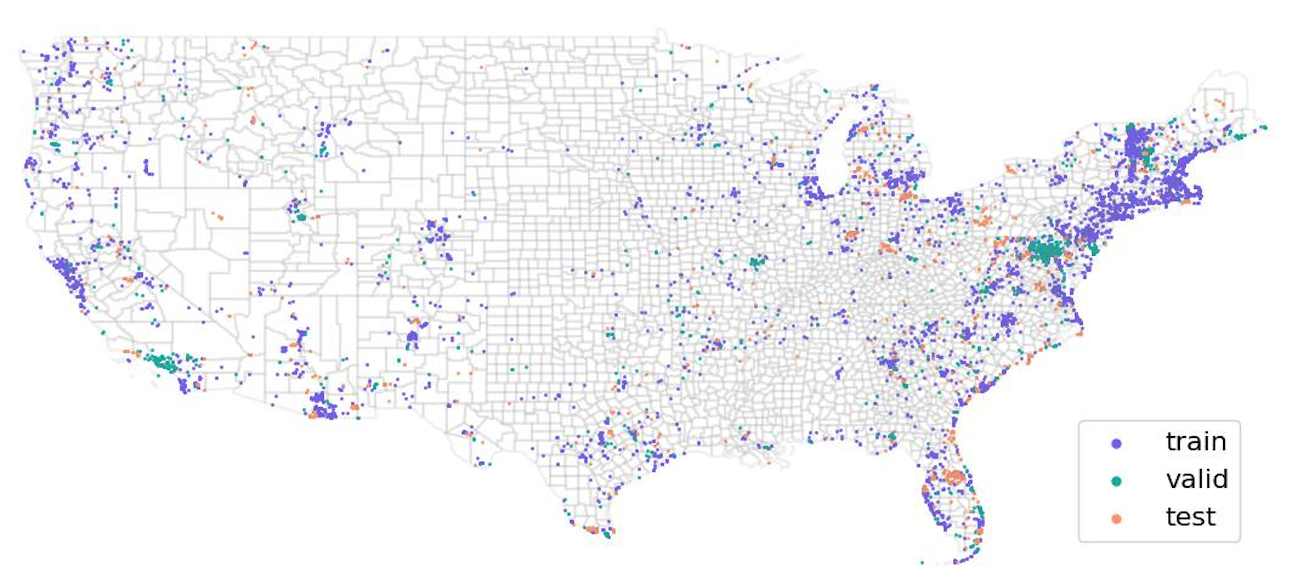}
    \caption{Geographic distribution of training, validation, and test splits in the SatButterfly-v2 dataset.}
    \label{fig:satbutterfly_v2_datadist}
\end{figure}

For readability and ease of use, we release the dataset in two versions: a) SatButterfly-v1, which includes locations across the USA, and b) SatButterfly-v2, which includes only locations that are co-located with those in SatBird. This alignment allows SatButterfly-v2 to be used for joint modeling of birds and butterflies, facilitating research on potential species interactions. The dataset is partitioned similarly to SatBird using the DBSCAN clustering algorithm from \verb|scikit-learn|  \citep{dbscan}. Table \ref{tab:num-samples-splits} summarizes the number of locations with records in each split for both versions of the SatButterfly dataset. Figures \ref{fig:satbutterfly_v1_datadist} and \ref{fig:satbutterfly_v2_datadist} show the geographic distribution of train/validation/test splits.

SatButterfly dataset is publicly available through this \href{https://huggingface.co/datasets/cisosdm/SatButterfly}{link}.

\begin{table}
    \begin{center}
    \renewcommand{\arraystretch}{1.1}
    \begin{tabular}{lcc}
        \hline
      & SatButterfly-v1 & SatButterfly-v2   \\ \hline
        train & $5316$  & $4677$ \\
        validation   & $1147$  & $1002$ \\
        test  & $1145$  & $1005$ \\
    \hline
    \end{tabular}
    \caption{Number of samples in each split for the two versions of SatButterfly.}
    \label{tab:num-samples-splits}
    \end{center}
\end{table}

\section{Experimental Setup: Additional Details and Results}
\label{appendix:exp_details}

In this section, we provide additional information on the model architectures and training procedures.
All models use a hidden dimension of $256$. An ablation study on the size of hidden dissension is conducted in Appendix \ref{appendix:ablation_hidden_dim}. For each experimental setup, we apply slightly different training configurations:
\begin{itemize}
    \item \textbf{Within-dataset sPlotOpen:} We use a learning rate of $0.001$ and a batch size of $64$, $n_b = 1$ for CISO, training all models for a maximum of $20$ epochs.
    \item \textbf{Within-dataset SatBird:} We use a learning rate of $0.0001$ and a batch size of $128$, training for up to 50 epochs.
    \item \textbf{Across-dataset SatBird and SatButterfly:} We use the same configuration as the within-dataset SatBird setup.
    \item \textbf{Across-dataset SatBird and sPlotOpen:} We use the same configuration as the within-dataset SatBird setup, except that we adjust the number of bins to $n_b=1$ for CISO, given that sPlotOpen labels are binary (presence/absence only).
\end{itemize}
We use the top-k metric on the validation set to keep the best model checkpoint, in the unconditioned setting, over all species. For the across-dataset experiments, the best checkpoint is selected based on improved metrics over both sets of species, computed with separate metrics (top-$k$ for SatBird, top-$k$ for SatButterfly, AUC for sPlotOpen). In other words, a model checkpoint is considered better than the previous one if it has higher metrics on both datasets A and B together. This guarantees that the best model achieves better results on the validation set for both dataset A and dataset B, without favoring one over the other.


\subsection{Additional metrics}
\label{appendix:add_metrics}

While we reported top-k and MAE results for SatBird in the main paper, Table \ref{tab:extra_within_dataset_satbird} presents additional numerical results for the MSE, top-10 accuracy, and top-30 accuracy metrics.
\begin{table}[ht]
    \renewcommand{\arraystretch}{1.2}
    \centering
    \resizebox{\columnwidth}{!}{%
        \begin{tabular}{l|cc|cccc|}
             & \multicolumn{6}{c|}{SatBird} \\ \hline
             & \multicolumn{2}{c|}{MSE $[\times 10^{2}]$} & \multicolumn{2}{c|}{Top-$10$ (\%)} & \multicolumn{2}{c|}{Top-$30$ (\%)} \\ \hline
            \multicolumn{1}{c|}{Model} & songbird & non-songbird & songbird & \multicolumn{1}{c|}{non-songbird} & songbird & non-songbird \\ \hline
            \textbf{Unconditioned} & & & & & & \\
            Linear & $1.56 \pm 0.00$ & $0.84 \pm 0.00$ & $45.24 \pm 0.03$ & \multicolumn{1}{c|}{$52.92 \pm 0.03$} & $70.03 \pm 0.04$ & $74.23 \pm 0.14$ \\
            Maxent & $1.09 \pm 0.00$ & $0.38 \pm 0.00$ & $50.25 \pm 0.01$ & \multicolumn{1}{c|}{$56.85 \pm 0.02$} & $76.06 \pm 0.04$ & $83.73 \pm 0.07$ \\
            MLP & $1.06 \pm 0.00$ & $0.38 \pm 0.00$ & $51.03 \pm 0.04$ & \multicolumn{1}{c|}{$57.89 \pm 0.11$} & $77.14 \pm 0.06$ & $84.71 \pm 0.14$ \\
            MLP++ & $1.12 \pm 0.01$ & $0.40 \pm 0.00$ & $47.76 \pm 0.45$ & \multicolumn{1}{c|}{$55.49 \pm 0.07$} & $75.01 \pm 0.14$ & $83.25 \pm 0.01$ \\
            CISO & $1.05 \pm 0.01$ & $0.37 \pm 0.00$ & $52.62 \pm 0.47$ & \multicolumn{1}{c|}{$59.28 \pm 0.33$} & $78.41 \pm 0.26$ & $85.78\pm 0.16$ \\ \hline
            \textbf{Conditioned} & & & & & & \\
            MLP++ & $0.86 \pm 0.01$ & $0.31 \pm 0.00$ & $55.11 \pm 0.28$ & \multicolumn{1}{c|}{$62.59 \pm 0.24$} & $78.92 \pm 0.28$ & $87.02 \pm 0.13$ \\
            CISO & $\mathbf{0.75} \pm 0.02$ & $\mathbf{0.28} \pm 0.00$ & $\mathbf{59.39} \pm 0.44$ & \multicolumn{1}{c|}{$\mathbf{65.4} \pm 0.42$} & $\mathbf{81.88} \pm 0.24$ & $\mathbf{88.59} \pm 0.35$
        \end{tabular}%
    }
    \setlength{\abovecaptionskip}{10pt}
    \caption{\textbf{Within the SatBird dataset}: Performance comparison of the different approaches applied to different species groups within the same dataset. Models with \textit{conditioned} inference incorporate partial information about the presence or absence of species from the other group. \textbf{Bolded} scores indicate the best performance for the given metric.}
    \label{tab:extra_within_dataset_satbird}
\end{table}

\section{Ablation Studies}

In this section, we describe and present further experiments that guided our final model design choices, based on ablation studies of several parameters.

\subsection{CISO: binning vs. continuous projection}
\label{appendix:ablation_continuous_projection}

In addition to discretizing encounter rates into four bins, we also experiment with linear and periodic encodings, following \cite{gorishniy2022embeddings}. We further evaluate a binary presence/absence encoding, which can be viewed as a special case of binning with $n_b = 1$. Results from this ablation study on the within-dataset SatBird task are reported in Table \ref{tab:extra_binning}.

\begin{table}[ht]
    \renewcommand{\arraystretch}{1}
    \centering
    \small
        \begin{tabular}{l|cc|cc|}
             & \multicolumn{4}{c|}{SatBird} \\ \hline
             & \multicolumn{2}{c|}{MAE $[\times 10^{2}]$} & \multicolumn{2}{c|}{Top-$k$ (\%)} \\ \hline
            \multicolumn{1}{c|}{Encoding strategy} & songbird & non-songbird & songbird & \multicolumn{1}{c|}{non-songbird}\\ \hline
            \textbf{Unconditioned} & & & & \\
            4 bins & 3.09 & 1.24  & 68.98&\multicolumn{1}{c|}{59.39} \\ 
            1 bin & 3.24 & 1.35 & 69.16& \multicolumn{1}{c|}{58.77}  \\
            Periodic & 3.13 & 1.27 & 69.39 & \multicolumn{1}{c|}{59.00} \\
            Linear & 3.19 & 1.29 &  69.33& \multicolumn{1}{c|}{59.19} 
            \\\hline
            \textbf{Conditioned} & & & & \\
            4 bins & \textbf{2.45} & \textbf{0.96} & \textbf{73.08}&\multicolumn{1}{c|}{\textbf{64.41}} \\ 
            1 bin & 2.75& 1.10 & 72.68 & \multicolumn{1}{c|}{63.39}  \\
            Periodic &2.47 & 1.02 & 72.76& \multicolumn{1}{c|}{63.44} \\
            Linear & 2.47 & 1.00 &72.98 & \multicolumn{1}{c|}{63.53} 
            \\\hline

        \end{tabular}%
    \setlength{\abovecaptionskip}{10pt}
    \caption{\textbf{Within the SatBird dataset}: Comparison of different encounter rate encoding strategies on the SatBird dataset. \textbf{Bolded} scores indicate the best performance for the given metric.}
    \label{tab:extra_binning}
\end{table}

\subsection{Number of layers for the MLP baseline} \label{appendix:ablation_num_layers}

We conduct an ablation study on the number of layers and parameters of the MLP baseline to verify if the model capacity is not a limiting factor of its performance. We use the sPlotOpen dataset for this ablation. We vary the number of layers and double the hidden dimension size as we add more layers, experimenting with 5, 6, and 7 layers, as opposed to the 3-layer MLP considered throughout the paper. Table \ref{tab:number_of_layers_ablation} shows results for an MLP model with 3, 5, 6, and 7 layers, while the number of parameters for the biggest model (7 layers) exceeds the number of parameters of CISO. We also show the results for MLP++ and CISO for reference. All models are trained under the same configuration set, for the same number of epochs.

\begin{table}[ht]
    \renewcommand{\arraystretch}{1.2}
    \centering
    \small
    {%
        \begin{tabular}{l|cc|cccc|}\\
             & \multicolumn{2}{c|}{AUC (\%)} \\ \hline
            \multicolumn{1}{c|}{} &  \hspace{9pt}tree\hspace{9pt} & non-tree & num of parameters \\ \hline
            \textbf{MLP: \#layers} & \multicolumn{1}{l}{} &  &  \\
            MLP-3 (baseline) & 98.42 & 96.56 & 1.1M \\
            MLP-5 & 98.32 & 96.69 & 3.6M \\
            MLP-6 & 98.52 & 96.76 & 5.4M \\
            MLP-7 & 98.44 & 96.63 & 7.8M \\ \hline
            \textbf{Unconditioned} & & &  \\
            MLP++ & 98.12 & 96.09 & 5.2M \\
            CISO & 98.52 & 96.55 & 7.1M \\ \hline
            \textbf{Conditioned} & & &  \\
            MLP++ & 98.78 & 96.89 & 5.2M \\
            CISO & \textbf{99.15} & \textbf{97.54} & 7.1M \\ \hline
           
        \end{tabular}%
    }
    \setlength{\abovecaptionskip}{10pt}
    \caption{\textbf{Ablation study on the number of layers for MLP baseline}: On the sPlotOpen dataset, we report results with the number of layers of MLP to 5, 6, or 7 compared to the baseline of 3. For reference, we also report the number of parameters for MLP++ and CISO along with their best performing results in the conditioned case.}
    \label{tab:number_of_layers_ablation}
\end{table}

\subsection{Size of hidden dimension} \label{appendix:ablation_hidden_dim}

We evaluate the effect of varying the hidden dimension size of the CISO model on the SatBird dataset, with results reported in Table \ref{tab:hid_dim_ablations}. Our main experiments used $256$ as the size of hidden dimension for all models.

\begin{table}[ht]
    \centering
    \small
        \begin{tabular}{l|cc|cc|}
             & \multicolumn{4}{c|}{SatBird} \\ \hline
             & \multicolumn{2}{c|}{MAE $[\times 10^{2}]$} & \multicolumn{2}{c|}{Top-$k$ (\%)} \\ \hline
            \multicolumn{1}{c|}{Hidden dimension} & songbird & non-songbird & songbird & \multicolumn{1}{c|}{non-songbird}\\ \hline
            \textbf{Unconditioned} & & & & \\
            64 & 3.38& 1.35  & 67.48&\multicolumn{1}{c|}{57.53} \\ 
            128 & 3.29 & 1.33 & 68.84& \multicolumn{1}{c|}{58.71}  \\
            256 & 3.10 & 1.24 & 68.98 & \multicolumn{1}{c|}{59.39} 
            \\\hline
            \textbf{Conditioned} & & & & \\
            64 & 2.69 & 1.18 & 71.20&\multicolumn{1}{c|}{60.17} \\ 
            128& 2.65& 1.03 & 72.33 & \multicolumn{1}{c|}{63.15}  \\
            256 &\textbf{2.45} & \textbf{0.97} & \textbf{73.08}& \multicolumn{1}{c|}{\textbf{64.41}} 
            \\\hline

        \end{tabular}%
    \setlength{\abovecaptionskip}{10pt}
    \caption{\textbf{Within the SatBird dataset}: Comparison of different sizes of hidden dimension for CISO on the SatBird dataset. \textbf{Bolded} scores indicate the best performance for the given metric.}
    \label{tab:hid_dim_ablations}
\end{table}



 
\end{appendix}

\end{document}